\documentclass[11pt]{article}
\usepackage{newtxtext,newtxmath}

\usepackage[final]{acl}

\usepackage{times}
\usepackage{latexsym}

\usepackage[T1]{fontenc}
\usepackage[utf8]{inputenc}

\usepackage{microtype}

\usepackage{inconsolata}

\usepackage{booktabs}     
\usepackage{multirow}      
\usepackage{threeparttable}
\usepackage{adjustbox}      
\usepackage[table]{xcolor} %

\usepackage[most]{tcolorbox}
\usepackage{enumitem}
\setlist[itemize]{nosep, leftmargin=1.5em, topsep=2pt, itemsep=1pt}
\usepackage{makecell}
\usepackage{graphicx}

\usepackage{etoolbox}
\usepackage{needspace}
\usepackage{makecell}

\usepackage{subcaption}

\renewcommand{\arraystretch}{0.9}
\tcbset{
	agentstyle/.style={
		enhanced,
		colback=blue!3!white,
		colframe=blue!40!black,
		arc=6pt,
		left=6pt,
		right=6pt,
		top=5pt,
		bottom=6pt,
		fonttitle=\bfseries\color{white},
	}
}
\usepackage{array}
\newcolumntype{L}[1]{>{\raggedright\arraybackslash}m{#1}}

\title{EvoSci: A Bio-Inspired Multi-Agent Framework for the Evolution of Scientific Discovery}

\author{
	\textbf{Xiaoyu Xiong},
	\textbf{Yuqi Ren\textsuperscript{$\dagger$}},
	\textbf{Deyi Xiong\textsuperscript{$\dagger$}}
	\\
	TJUNLP Lab, School of Computer Science and Technology, Tianjin University, China
	\\
	\{2025244184, ryq20, dyxiong\}@tju.edu.cn
}
\begin{document}
\maketitle
\begingroup
\renewcommand\thefootnote{}
\footnotetext{\textsuperscript{$\dagger$} Corresponding authors.}
\endgroup

%\AtBeginEnvironment{equation}{%
%	\setlength{\abovedisplayskip}{6pt}%
%	\setlength{\belowdisplayskip}{6pt}%
%	\setlength{\abovedisplayshortskip}{4pt}%
%	\setlength{\belowdisplayshortskip}{4pt}%
%}
\setlength{\abovedisplayskip}{6pt}%
\setlength{\belowdisplayskip}{6pt}%
\setlength{\abovedisplayshortskip}{4pt}%
\setlength{\belowdisplayshortskip}{4pt}%

\begin{abstract}
	Large language models (LLMs), have shown strong potential in scientific discovery, yet existing methods still face substantial challenges in the design of research workflows and multi-role collaboration mechanisms. To mitigate these issues, we propose EvoSci, a multi-agent scientific collaboration framework, which integrates bio-inspired evolution with knowledge graph modeling. To iteratively generate, evaluate, and refine research ideas, EvoSci incorporates multiple role-based agents, including mentor, researcher, and reviewer. By combining collaborative reasoning, shared memory, and evolutionary feedback, EvoSci significantly enhances the coherence and creativity of scientific exploration. Experiments on real-world research topics demonstrate that EvoSci significantly outperforms strong baselines in LLM-based structured peer-review and comparative ranking evaluations, achieving the highest overall peer-review score (ICLR 4.90) and top ranking (Top-10 = 54). These results suggest its superiority in both scientific idea generation and continuous discovery.
\end{abstract}

\section{Introduction}
With the breakthrough development in knowledge representation \cite{pan2024unifying}, logical reasoning \cite{ke2025survey,xu2025towards}, and complex multimodal information integration \cite{han2025surveygenerativecategoriestechniques}, LLMs are gradually reshaping the paradigm of scientific research \cite{buehler2024accelerating}. Significant progress has already been made in traditional AI-powered domains such as mathematical reasoning and theorem proving \cite{trinh2024solving,zhang-xiong-2025-debate4math,liu-etal-2025-highmath}, automatic code generation \cite{li2023skcoder,ren2023misuse,he-etal-2024-cocost,Yang2025ProBenchBL}, and complex data analysis \cite{sui2024table}. Building on these successes, researchers are increasingly exploring the potential of LLMs across broader scientific workflows, including idea and hypothesis generation from large-scale scientific corpora \cite{kulkarni2025scientifichypothesisgenerationvalidation,zhou-etal-2024-hypothesis,wang2024scipip,yang2024large,Wang_2024}, experimental design \cite{desai2025autoscilab,tian2021autooedautomatedoptimalexperiment,noh2024integrated,tom2024self}, and result interpretation \cite{zheng2023large,Charness2025}.

However, scientific discovery is not a one-shot solution but a gradual and evolving process, driven by the continual refinement of research problems and the accumulation of intermediate insights \cite{ELLIOTT2012376}. It is also fundamentally collaborative, relying on the interplay of diverse roles and perspectives \cite{doi:10.1073/pnas.1309723111}. However, existing approaches often reduce LLMs to static executors within rigid pipelines, overlooking their potential for long-horizon inquiry and structured coordination. This raises two central challenges: (1) how could LLMs be steered toward progressively deepening scientific problems? and (2) how could effective collaboration frameworks be developed to allow multiple agents to engage in sustained, dynamic exploration.

%To address these challenges, we propose \textbf{EvoSci}, an \textbf{Evo}lutionary \textbf{Sci}ence framework driven by multiple collaborative agents for automatic research ideation, consisting of four stages: \textbf{Problem Space Construction}, \textbf{Collaborative Research Execution}, \textbf{Research Idea Evaluation}, and\textbf{ Bio-Inspired Evolutionary Iteration}. EvoSci builds upon explicitly defined role-based agents, including a mentor, a group of researchers, and a reviewer, each responsible for a distinct stage of the ideation process. Beyond static, pre-defined pipelines, EvoSci enables adaptive coordination through dynamic task decomposition, where the mentor agent reallocates subtasks based on intermediate feedback, while role-aware assignment ensures that each agent's actions remain aligned with its disciplinary background across multiple interaction rounds. Furthermore, inspired by biological evolution, we iteratively update research ideas by aligning and recombining conceptual knowledge across different domains, thereby enhancing the novelty of scientific exploration.

To address these challenges, we propose \textbf{EvoSci}, an \textbf{Evo}lutionary \textbf{Sci}ence framework driven by multiple collaborative agents for automatic research ideation. Inspired by real-world research teams and biological evolution, EvoSci models scientific discovery as a long-horizon, iterative exploration, consisting of four stages: \textbf{Problem Space Construction}, \textbf{Collaborative Research Execution}, \textbf{Research Idea Evaluation}, and \textbf{Bio-Inspired Evolutionary Iteration}. EvoSci builds upon explicitly defined role-based agents, including a mentor, a group of researchers, and a reviewer, each responsible for a distinct stage of the ideation process. Beyond static, pre-defined pipelines, EvoSci enables adaptive coordination through dynamic task decomposition, where the mentor agent reallocates subtasks based on intermediate feedback, while role-aware assignment ensures that each agent's actions remain aligned with its disciplinary background across multiple interaction rounds. Furthermore, inspired by biological evolution, we iteratively update research ideas by aligning and recombining conceptual knowledge across different domains, thereby enhancing the novelty of scientific exploration.

To validate the effectiveness and generality of EvoSci, we have conducted systematic experiments across diverse scientific scenarios. Results show that EvoSci consistently generates more novel and impactful research ideas than strong baselines. Equipped with DeepSeek-v3, EvoSci achieves the highest overall peer-review scores (ICLR 4.90 / NeurIPS 3.95), surpassing the next best baseline (4.68 / 3.72) by a large margin, and maintaining consistent advantages in terms of Elo-based ranking metrics (Avg Wins 4.19, Top-10 Count 47). These results validate that EvoSci achieves superior overall research quality and more reliable relative performance compared to strong baselines. In summary, the main contributions of our work are as follows:
\begin{itemize}
	\item We conceptualize scientific discovery as a problem-oriented process, in which research problems are dynamically generated and progressively refined through a multi-agent collaboration loop.
	\item We construct a heterogeneous multi-agent framework that mirrors real-world research laboratories, where diverse agents operate under role-specific objectives. The framework is grounded on datasets derived from real scientists, enabling more authentic simulation of collaborative scientific workflows.
	\item We implement a multi-round feedback with bio-inspired evolution mechanism (selection, crossover, mutation) to enable continuous and open-ended scientific exploration.
\end{itemize}

\section{Related Work}
Our work is related to both AI-driven scientific discovery and multi-agent systems. We briefly review these two topics within the scope of LLM and the constraint of space. 
\subsection{AI for Scientific Discovery}
Recent advances in LLMs have enabled AI systems to participate more deeply in scientific discovery \cite{zheng2025automation,doi:10.36227/techrxiv.177155935.57684125/v1}. Early systems primarily assist with literature mining \cite{SMALHEISER1998149,HRISTOVSKI2005289} and experiment design \cite{doi:10.1126/science.1165620,tian2021autooedautomatedoptimalexperiment}, while recent developments aim at more comprehensive support for the scientific process. Notably, SciPIP employs three retrieval strategies (semantic, entity, and co-occurrence) to enhance hypothesis generation~\cite{wang2024scipip}, and the MOOSE framework introduces iterative feedback mechanisms to evolve hypotheses from large-scale web corpora~\cite{yang2024large}. Building on these advances, SciAgents integrates multi-agent reasoning with scientific knowledge graphs to autonomously generate, test, and refine hypotheses~\cite{ghafarollahi2025sciagents}, while CoScientist further extends such capabilities by autonomously planning and executing experimental procedures within chemical research workflows~\cite{jansen2025codescientist}. 

Recent systems have taken a more ambitious step toward end-to-end scientific discovery. For example, AI-Scientist \cite{lu2024ai,yamada2025ai} and CycleResearcher \cite{weng2024cycleresearcher} automate nearly the entire scientific workflow, from topic formulation and literature exploration to hypothesis generation, experiment simulation, paper writing, and even peer review. However, existing approaches primarily focus on conducting a single round of research under a fixed initial topic, overlooking the evolutionary and cyclic nature of scientific discovery. This gap highlights the need for frameworks that support iterative refinement and problem reformulation across successive research cycles.
﻿
\begin{figure*}[t]
	\centering
	\includegraphics[width=0.95\linewidth]{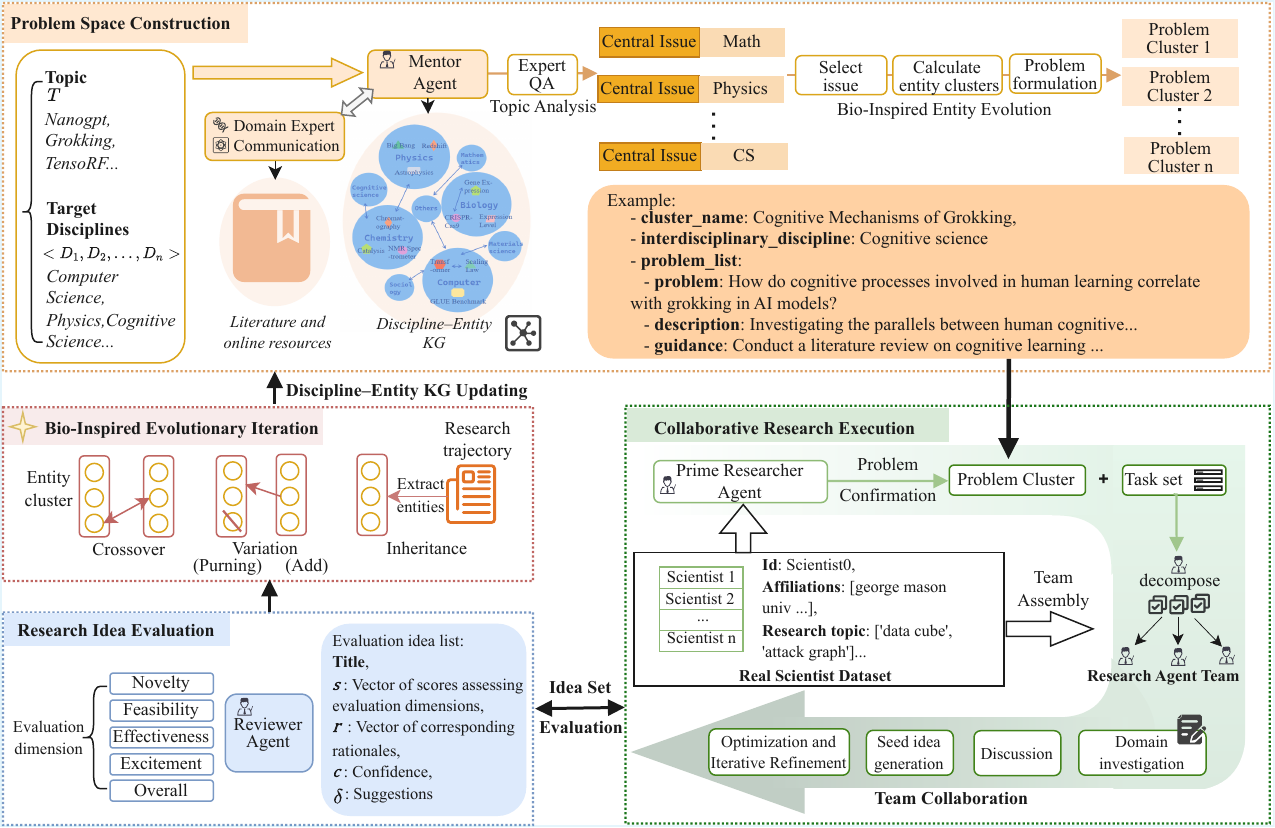}
	\vspace{-5pt}
	\caption{Overall workflow of the proposed EvoSci framework. EvoSci begins with problem space construction from literature and domain knowledge, followed by collaborative research execution through role-based agents and iterative evaluation with reviewer feedback. The bio-inspired evolutionary loop operates over multiple rounds, leveraging feedback to recombine, adapt, and refine research directions.}
	\label{fig:workflow}
\end{figure*}

\subsection{Collaboration in Multi-Agent Systems}
%With the rapid advancement of large language models, the application boundaries of artificial intelligence have been continuously expanding. LLMs demonstrate the ability to handle complex tasks, in which their performance in some cases approaches or even surpasses that of humans. Nevertheless, despite their remarkable potential, single LLMs still exhibit notable limitations in long-term reasoning, self-reflection, and cognitive consistency, manifesting issues such as hallucination~\cite{huang2025survey}, limited long-horizon reasoning~\cite{ferrag2025reasoninglimitsadvancesopen}, and stale or outdated knowledge~\cite{zhang2023large}.
%
%To overcome these problems, the emerging paradigm of \emph{agentic artificial intelligence} (Agentic AI) introduces the notions of \emph{actionability} and \emph{environmental interaction}. In this paradigm, the LLM serves as the “cognitive core” or “central reasoning engine,” complemented by external tools (e.g., code executors, databases, retrievers) and task-oriented modules (e.g., planners, controllers) to construct executable agents~\cite{durante2024agent}. This architecture enables AI systems to interact more realistically with their environments and has yielded new solutions for complex domains such as scientific discovery~\cite{li2024chain}, programming~\cite{huang2023agentcoder}, education~\cite{viswanathan2022enhancement}, and healthcare~\cite{li2024agent}. 
Recent work has increasingly turned to multi-agent systems (MAS) as a way to mitigate the limitations of single LLMs, such as hallucination~\cite{huang2025survey}, weak long-horizon reasoning~\cite{ferrag2025reasoninglimitsadvancesopen}, and stale knowledge~\cite{zhang2023large}. MAS build on the paradigm of agentic AI~\cite{durante2024agent}, organizing multiple LLM-based agents with specialized roles and shared goals. Through coordinated planning, division of labor, and mutual evaluation, these systems aim to achieve more reliable and scalable cognitive performance than any individual model~\cite{NEURIPS2024_0852b88e,li-etal-2025-chatsop}.

A growing line of studies applies MAS to scientific discovery, demonstrating their potential in iterative and multi-step research workflows. VIRSCI simulates a virtual team of scientists engaging in structured idea generation and evaluation~\cite{su-etal-2025-many-copy}, while ResearchAgent coordinates specialized agents for literature analysis, hypothesis generation, and experiment planning~\cite{baek2024researchagent}. These works largely remain task- or pipeline-oriented, and fall short of providing an integrated framework that supports cyclic scientific evolution together with role-aware, interdisciplinary collaboration.
\section{EvoSci}
EvoSci models scientific discovery as an evolutionary, problem-centric process, where research directions are iteratively explored, evaluated, and refined over long horizons. Accordingly, EvoSci consists of four core components: (1) a problem space construction module guided by a mentor agent, (2) a collaborative research execution module led by a prime researcher agent, (3) an evaluation module that systematically assesses the quality, novelty, and feasibility of generated research ideas, and (4) a bio-inspired evolutionary iteration module that enables iterative refinement of research directions. Together, these components form a closed-loop workflow for the iterative generation and refinement of interdisciplinary research ideas. Figure~\ref{fig:workflow} illustrates the overall workflow of EvoSci.

\subsection{Problem Space Construction}
A core challenge in automating scientific idea generation lies in the construction of a structured, high-quality problem space that supports exploration across disciplinary boundaries. This phase aims to transform an initial research theme into a diverse collection of research problems that are both semantically grounded and structurally expandable.

\noindent\textbf{Data Preparation.}
We construct a lightweight, multi-level knowledge graph to organize scientific disciplines and their associated entities.
We begin with a predefined set of representative disciplines spanning major scientific disciplines (e.g., Physics, Chemistry, Biology, Medicine, Economics), each represented as a first-layer node.

For each discipline, we extract candidate entities from its Wikipedia page using the page summary and hyperlink structure.
An LLM-based classifier assigns each entity a semantic type (e.g., \emph{Theory}, \emph{Model}, \emph{Material}, \emph{Phenomenon}) and estimates its relevance to the discipline. Only relevant entities are retained and connected to their corresponding disciplines via \emph{has\_entity} edges.

To capture cross-disciplinary connections, we compute cosine similarity between the embedding representations of entities and add a cross-entity edge if the similarity exceeds a threshold, i.e., $s(e_i,e_j)=\cos(\mathrm{emb}(e_i),\mathrm{emb}(e_j))>\tau$.

\noindent\textbf{Topic Analysis.}
Given a core research topic $T$ and a set of target disciplines $\mathcal{D}_{\mathrm{target}}=\{D_1,\dots,D_n\}$, the system first grounds the topic by using an LLM-based classifier to map $T$ to one or more core disciplines in the knowledge graph, ensuring that subsequent exploration is anchored in a clear scientific context. To encourage cross-disciplinary integration, we adopt a question-answering architecture with domain expert agents instantiated from a dataset of real-world scientists (Appendix~\ref{appendix:datasetx}). Each expert agent is equipped with literature retrieval and reading capabilities and engages in structured discussions with a mentor agent. Through this process, the system identifies promising interdisciplinary directions and updates the knowledge graph with domain-specific entities derived from their exploration trajectories, enabling iterative evolution to support downstream idea generation.
﻿

\noindent\textbf{Bio-Inspired Entity Evolution.}
After the topic analysis concludes, the system enters the problem generation stage by focusing on the intersection between the topic $T$ and a selected discipline $d \in \mathcal{D}_{\mathrm{target}}$. For discipline $d$, $\mathcal{E}_d$ denotes the set of associated entities in the knowledge graph, which are clustered into semantic entity clusters $\mathcal{C}_d=\{\mathcal{C}_{d,1},\ldots,\mathcal{C}_{d,n}\}$. The most topic-relevant cluster is then explicitly incorporated into the prompt of the mentor agent, which uses these entity clusters as contextual cues to generate scientific research problems:
\begin{equation}
	Q_\mathrm{{problem}} \;=\; \langle\, T,\; d,\; \mathrm{Top}(\mathcal{C}_d;\,T)\,\rangle ,
	\label{eq:prompt_cluster}
\end{equation}
where $\mathrm{Top}(\mathcal{C}_d;T)$ denotes the single entity cluster in $\mathcal{C}_d$ that is most semantically relevant to the topic $T$. After each exploration round, bio-inspired evolutionary operations (i.e., \emph{Crossover}, \emph{Variation}, \emph{Inheritance}, and \emph{Selection}) are applied at the cluster level to the discipline entity space, as detailed in Section~\ref{sec:evaluation_feedback}.

\noindent\textbf{Structured Problem Cluster Generation.}
Driven by LLMs, the system generates a diverse set of candidate scientific problems $\mathcal{Q}=\{q_1,q_2,\dots\}$ by expanding along the evolving entity clusters in the knowledge graph. Each problem $q_i$ is represented by a concise problem statement $\mathcal{P}(q_i)$, an explanatory description $\mathcal{D}(q_i)$, and a research guidance field $\mathcal{G}(q_i)$.

The generated problems are subsequently grouped into problem clusters $\mathcal{P}=\{\mathcal{P}_1,\mathcal{P}_2,\dots\}$ according to their primary interdisciplinary foci, forming structured problem sets that emphasize cross-disciplinary perspectives and conceptual novelty.

﻿

%\subsection{Collaborative Research Execution Module: Multi-Agent Collaboration-Driven Scientific Process}
\subsection{Collaborative Research Execution}
Building on the structured problem clusters, the system proceeds to a multi-agent research exploration phase.  A set of role-specialized agents collaboratively investigates selected problem clusters through literature review, structured discussion, and iterative idea generation and refinement, simulating the workflow of real scientific teams.

\noindent\textbf{Problem Confirmation and Team Assembly.}
From the candidate problem clusters $\mathcal{P}$, the system selects a target cluster $\mathcal{P}^*$ by jointly considering its relevance to the initial topic, interdisciplinary potential, and future extensibility. 
A research team is then assembled to explore $\mathcal{P}^*$. The prime researcher and assistant researchers are instantiated from a real-world scientist dataset, where each agent is represented by anonymized metadata and semantic behavior embeddings. The prime researcher is selected to align with the initial topic, while assistant researchers are chosen based on their relevance to $\mathcal{P}^*$, facilitating effective interdisciplinary collaboration.

\noindent\textbf{Task Decomposition and Team Collaboration.}
Given a selected problem cluster $P^*$, the Prime Researcher organizes a collaborative research process by decomposing the exploration into a set of structured tasks $\mathcal{T}=\{t_1,t_2,\dots\}$. These tasks correspond to key stages of scientific inquiry, including background investigation, problem analysis, idea generation, and iterative refinement. We adopt the CrewAI framework to organize the agent team and establish a dynamic delegation mechanism based on a “lead-and-collaborate” interaction paradigm, which supports the following collaboration processes:
﻿
\begin{itemize}
	\item \textbf{Task Leading and Delegation:} 
	Each core task is initiated and led by the Prime Researcher, who analyzes the task context, decomposes it into subtasks, and assigns them to assistant agents based on semantic relevance and skill profiles.
	
	\item \textbf{Recursive Delegation and Collaboration:} 
	Assistant agents may further decompose assigned subtasks and reassign them when necessary, resulting in a recursive collaboration structure.
	
	\item \textbf{Phased Integration:} 
	At predefined stages of task execution, structured discussion rounds are conducted to aggregate intermediate results, align perspectives across agents, and integrate subtask outcomes.
\end{itemize}

Formally, for a task $t_k$, we define the research state and prompt as:
\begin{equation}
	\mathcal{S}_k
	=
	\langle\,
	t_{\mathrm{description}},\;
	\bigcup_{i=1}^{k-1}\mathcal{R}_i,\;
	\mathcal{R}_k^{\mathrm{sub}},\;
	a_k
	\,\rangle ,
	\label{eq:exploration_state}
\end{equation}
\begin{equation}
	Q_{\mathrm{research}}
	=\langle\,
	T,\; P^*,\; \mathcal{S}_k
	\,\rangle ,
	\label{eq:task_prompt}
\end{equation}
where $t_{\mathrm{description}}$ denotes the description of the current task or subtask under exploration, $\bigcup_{i=1}^{k-1}\mathcal{R}_i$ represents the aggregated responses accumulated from previously completed tasks,
$\mathcal{R}_k^{\mathrm{sub}}$ denotes the intermediate responses produced for subtasks under the current task $t_k$,
and $a_k$ denotes the agent assigned to execute $t_k$.
The research prompt $Q_{\mathrm{research}}$ combines the initial topic $T$, the selected problem cluster $P^*$, and the current research state $\mathcal{S}_k$ to guide the task execution. In practice, the accumulated responses in $\mathcal{S}_k$ are maintained through a hierarchical memory mechanism, including short-term memory, long-term memory, and entity memory, which enables compact context representation and more precise summarization across tasks.
﻿
%This mechanism enhances the Prime Researcher's control over the research process while fostering active knowledge exchange and task evolution among agents, establishing a cognitive basis for interdisciplinary collaboration in complex scientific tasks.

\noindent\textbf{Seed Idea Generation and Refinement.}
Through domain investigation and collaborative discussion, the system generates diverse seed ideas, broadening the exploration space and increasing the likelihood of high-quality discoveries. These ideas are refined through multi-agent collaboration, where redundant or low-value ideas are removed, the remaining ideas are ranked by novelty, feasibility, and interdisciplinary value.

%To further refine these proposals, a reviewer agent is engaged to simulate peer review. Each refined idea $\mathcal{I}_{refined}$ is evaluated across dimensions, including novelty, feasibility, validity , and scientific excitement. The reviewer agent provides critical feedback, identifies logical weaknesses, suggests complementary experimental directions, and delivers quantitative scores along with rationale. This structured evaluation introduces external perspectives to prevent groupthink and strengthen the rigor and innovativeness of the final outputs.
%
%Furthermore, to maintain consistency across iterations, a hierarchical memory mechanism is introduced. It includes short-term memory for local context, long-term storage for accumulated knowledge, and entity memory for tracking key concepts. These components help sustain coherent knowledge development and support interdisciplinary collaboration.

%\subsection{Evaluation-Feedback Loop: Evaluation-Guided Evolution for Long-Term Discovery}
\subsection{Research Idea Evaluation}
To further refine generated ideas, a reviewer agent is engaged to simulate a peer-review process. 
Each generated idea $\mathcal{I}=\{I_1,I_2,\dots\}$ is evaluated along multiple dimensions, including novelty, feasibility, validity, and scientific excitement. 
In addition to quantitative scores, the reviewer agent provides structured feedback by identifying logical weaknesses and suggesting complementary experimental directions.

Formally, the evaluation results can be expressed as:
\begin{equation}
	E(\mathcal{I})
	=\langle\,
	\mathrm{title},\;
	s,\;
	r,\;
	c,\;
	\delta
	\,\rangle ,
	\label{eq:evaluation}
\end{equation}
where $\mathrm{title}$ denotes the title of the evaluated idea,
$s = (s_{\mathrm{nov}}, s_{\mathrm{fea}}, s_{\mathrm{eff}}, s_{\mathrm{exc}}, s_{\mathrm{overall}})$ is a vector of scores assessing novelty, feasibility, expected effectiveness, scientific excitement and overall,
and $r = (r_{\mathrm{nov}}, r_{\mathrm{fea}}, r_{\mathrm{eff}}, r_{\mathrm{exc}}, r_{\mathrm{overall}})$ denotes the corresponding rationales.
$c$ denotes the reviewer's confidence level,
and $\delta$ denotes concrete suggestions for improving the idea.

\subsection{Bio-Inspired Evolutionary Iteration}
\label{sec:evaluation_feedback}
Scientific idea generation is inherently iterative rather than one-shot.
To support long-term and open-ended discovery, EvoSci introduces a bio-inspired evolutionary loop that operates on the entity layer of the knowledge graph and is guided by structured evaluation feedback.

\noindent\textbf{Entity-Level Evolution.}
Building on the multi-level knowledge graph, the discipline layer remains static as a structural backbone, while the entity layer is treated as an evolving population.
For each discipline, entities are organized into semantic clusters, which serve as the basic units of evolution.
Across successive exploration rounds, these entity clusters are iteratively updated to refine cross-disciplinary exploration.

Concretely, the system applies a set of bio-inspired evolutionary operations at the cluster level.
\begin{itemize}
	\item \textbf{Crossover:} 
	Enables recombination by exchanging entities between different semantic clusters within the same discipline, producing novel concept combinations while preserving domain coherence.
	\item \textbf{Variation:} 
	Injects diversity by introducing new or low-frequency entities into existing clusters, preventing premature convergence.
	
	\item \textbf{Selection:} 
	Filters entity clusters based on evaluation feedback from generated ideas, favoring clusters that exhibit higher novelty, feasibility, and relevance to the research topic.
	
	\item \textbf{Inheritance:} 
	Propagates high-fitness entities and clusters into subsequent iterations, ensuring that valuable knowledge accumulates over time.
\end{itemize}

\noindent\textbf{Evaluation-Guided Loop.}
After each exploration round, refined ideas and their evaluations are summarized by the Prime Researcher and passed to the Mentor Agent.
High-fitness entities identified from successful ideas are re-integrated into the knowledge graph, while low-value entities are pruned.
The updated entity clusters then serve as seeds for reconstructing the next problem set, forming a closed evolutionary loop that balances exploration and consolidation.

%Figure~\ref{fig:evolutionary_loop} illustrates the full research cycle, from initial problem generation to iterative refinement through evaluation and entity-level evolution.
﻿
%\begin{figure*}[t]
%	\centering
%	\includegraphics[width=0.8\textwidth]{./figure/evolutionary_loop.png}
%	\caption{The Evolutionary Loop of Multi-Agent Scientific Discovery}
%	\label{fig:evolutionary_loop}
%\end{figure*}

\section{Experiments}
%We designed the experimental setup to comprehensively examine EvoSci's adaptability across diverse scientific domains. Ten representative and challenging open research topics were selected, adopted from the task settings introduced in AI Scientist \cite{lu2024ai}, including 2D diffusion modeling, character-level language model training, earthquake prediction, neural network grokking behavior, NanoGPT, Materials Adaptive Convolutional Equivariants Model, SEIR infection modeling, sketch generation with recurrent neural networks, optimization of small CNN architectures with multi-dataset experiments, and neural rendering with TensoRF. For each task, the system was initialized with a simple topic prompt and then proceeded through problem-space construction, research exploration, evaluation feedback, and multi-round iteration.
%We designed an experimental setup to comprehensively examine EvoSci's adaptability across diverse scientific domains. Ten representative and challenging open research topics were selected from the task settings introduced in AI Scientist \cite{lu2024ai}. Our detailed task settings are provided in Appendix~\ref{appendix:task}, and the corresponding prompts used in the experiments are provided in Appendix~\ref{appendix:prompt}.
We designed an experimental setup to examine EvoSci's adaptability across diverse scientific domains. 
Ten representative and challenging open research topics were selected from the task settings introduced in AI Scientist \cite{lu2024ai}. 
Detailed task settings and corresponding experimental prompts are provided in Appendices~\ref{appendix:task} and \ref{appendix:prompt}.

\subsection{Evaluation Methodologies}
To comprehensively assess the effectiveness and creativity of our multi-agent scientific system, we adopted both qualitative and quantitative evaluations, combining an expert-simulated review mechanism with a tournament-style idea ranking procedure. In addition, we conducted an additional experiment to further validate the effectiveness and stability of the proposed meta-review mechanism (see Appendix~\ref{appendix:meta-review-validation}).

\noindent\textbf{Multi-Reviewer + Meta-Reviewer Mechanism.}  
Inspired by the AI Scientist evaluation framework \cite{lu2024ai}, we designed a structured LLM-driven peer-review workflow that simulates academic conference reviewing. Reviewer agents independently assess generated ideas using prompts aligned with ICLR and NeurIPS review templates, each incorporating a reflection mechanism to refine their evaluations. A meta-reviewer agent then aggregates individual reviews into a unified meta-review, enabling interpretable, reproducible, and academically representative evaluation.

\noindent\textbf{Tournament-Style Idea Ranking.}  
In addition, we implemented a comparative ranking procedure to evaluate idea quality. All generated ideas were pooled, randomized, and initially assigned one point. Ideas were paired and compared using the prompt: \emph{``One of them is accepted by a top AI conference (like ICLR or ACL) and the other one is rejected.''} The winner of each comparison received one point. This process was repeated for five rounds, and the final ranking was determined by aggregated scores. This tournament-style evaluation provided a robust relative measure of idea quality, complementing the structured review mechanism. Prior studies have also shown that such pairwise comparison, rather than absolute scoring, enables LLM-based evaluations to better align with human expert judgments.
﻿
\subsection{Main Results}
To systematically verify the overall effectiveness of EvoSci, we conducted experiments on the ten representative research topics described above. Our system was compared against four baseline methods: SciPIP, AI Scientist, COI Agent and VirSci (detailed descriptions are provided in the Appendix~\ref{appendix:baselines}). For fairness, each method was configured to generate 10 research ideas per topic, following comparable settings on iteration rounds, team size, and evaluation feedback. The generated ideas were evaluated using two mechanisms: an expert-simulated review and a tournament-style ranking.
The aggregated results are reported in Table~\ref{tab:main_results}, while the tournament-style ranking results are reported in Table~\ref{tab:agent_avgwin_top10count}.

\begin{table*}[ht]
	\centering
	\footnotesize
	\caption{Subjective evaluation scores of various agent models and methods.}
	\label{tab:main_results}
	\begin{threeparttable}
		\begin{adjustbox}{max width=0.9\textwidth}
			\footnotesize
			\begin{tabular}{llcccccc}
				\toprule
				\textbf{Agent Model} & \textbf{Method}
				& \textbf{Novelty} & \textbf{Feasibility}
				& \textbf{Validity} & \textbf{Excitement}
				& \textbf{ICLR Overall} & \textbf{NeurIPS Overall} \\
				\midrule
				
				\multirow{5}{*}{GPT-4o} 
				& AI Scientist & 4.31 & \textbf{6.49} & 4.87 & 4.11 & 4.33 & 3.19 \\
				\rowcolor{gray!10}
				& SciPIP & 4.72 & 4.76 & 4.74 & 4.49 & 4.26 & 3.06 \\
				& VirSci & \textbf{5.12} & 3.64 & 4.56 & \textbf{4.95} & 4.28 & 3.26 \\
				\rowcolor{gray!10}
				& CoI-Agent & 4.52 & 4.77 & 4.79 & 4.33 & 4.20 & 3.21\\
				& \textbf{EvoSci} & 4.78 & 4.62 & \textbf{5.01} & 4.75 & \textbf{4.45} & \textbf{3.44} \\
				
				\midrule
				\multirow{5}{*}{DeepSeek-v3} 
				& AI Scientist & 4.60 & \textbf{6.85} & 5.00 & 4.29 & 4.68 & 3.39 \\
				\rowcolor{gray!10}
				& SciPIP & 4.42 & 5.80 & 4.85 & 4.13 & 4.34 & 3.02 \\
				& VirSci & 5.48 & 3.95 & 4.88 & 5.11 & 4.50 & 3.69 \\
				\rowcolor{gray!10}
				& CoI-Agent & 5.07 & 4.66 & 4.92 & 4.58 & 4.51 & 3.72\\ 
				& \textbf{EvoSci} & \textbf{5.71} & 4.68 & \textbf{5.25} & \textbf{5.15} & \textbf{4.90} & \textbf{3.95} \\
				
				\midrule
				\multirow{5}{*}{Qwen3-max} 
				& AI Scientist & 4.18 & \textbf{7.00} & 5.04 & 4.15 & 4.48 & 3.31 \\
				\rowcolor{gray!10}
				& SciPIP & 4.87 & 5.41 & 4.89 & 4.53 & 4.54 & 3.17 \\
				& VirSci & \textbf{5.37} & 4.01 & 4.90 & \textbf{5.00} & 4.35 & 3.57 \\
				\rowcolor{gray!10}
				& CoI-Agent & 4.64 & 4.38 & 4.76 & 4.40 & 4.19 & 3.62\\ 
				& \textbf{EvoSci} & 5.14 & 4.98 & \textbf{5.20} & 4.89 & \textbf{4.72} & \textbf{3.81} \\
				\bottomrule
			\end{tabular}
		\end{adjustbox}
	\end{threeparttable}
\end{table*}

%In addition to the expert-simulated review, we adopted a tournament-style ranking to assess the relative preference of generated ideas through direct comparisons. This complementary evaluation focuses on pairwise wins and the frequency of being ranked among the top-performing ideas, reflecting the competitiveness of each method in head-to-head settings. The aggregated results of this evaluation are reported in Table~\ref{tab:agent_avgwin_top10count}.

\begin{table}[ht]
	\centering
	\small
	\caption{Average wins and top 10 counts for various agent models and methods.}
	\label{tab:agent_avgwin_top10count}
	\begin{threeparttable}
		\begin{adjustbox}{max width=0.4\textwidth}
			\footnotesize
			\begin{tabular}{llcc}
				\toprule
				\textbf{Agent Model} & \textbf{Method} & \textbf{Avg Wins} & \textbf{Top 10 Count} \\
				\midrule
				
				\multirow{5}{*}{GPT-4o} 
				& AI Scientist & 3.88 & 13 \\
				\rowcolor{gray!10}
				& SciPIP     & 2.70 & 7 \\
				& VirSci     & 4.07 & 52 \\
				\rowcolor{gray!10}
				& CoI-Agent & 3.58 & 36 \\
				& \textbf{EvoSci}    & \textbf{4.27}  & \textbf{54} \\
				
				\midrule
				
				\multirow{5}{*}{DeepSeek-v3} 
				& AI Scientist & 2.83 & 8 \\
				\rowcolor{gray!10}
				& SciPIP     & 2.50 & 3 \\
				& VirSci     & 3.90 & 35 \\
				\rowcolor{gray!10}
				& CoI-Agent & 4.08 & 37 \\
				& \textbf{EvoSci}    & \textbf{4.19}  & \textbf{47} \\
				
				\midrule
				
				\multirow{5}{*}{Qwen3-max} 
				& AI Scientist & 2.92 & 7 \\
				\rowcolor{gray!10}
				& SciPIP     & 2.39 & 1 \\
				& VirSci     & 3.94 & 34 \\
				\rowcolor{gray!10}
				& CoI-Agent & 4.00 & 39 \\
				& \textbf{EvoSci}    & \textbf{4.25}  & \textbf{50} \\
				
				\bottomrule
			\end{tabular}
		\end{adjustbox}
	\end{threeparttable}
\end{table}

From Table ~\ref{tab:main_results}, we observe that EvoSci consistently outperforms all baselines across models and evaluation dimensions. It achieves the highest scores on \textit{Validity}, \textit{Excitement}, and both overall metrics, indicating that the generated research ideas are not only credible but also engaging. While baselines such as AI Scientist or VirSci exhibit isolated strengths on individual criteria, their overall performance remains uneven, whereas our framework provides balanced improvements across all aspects. On the two overall measures, EvoSci yields 5--10\% relative gains in ICLR Overall (e.g., 4.90 vs.~4.68 with DeepSeek-v3, 4.72 vs.~4.54 with Qwen3-max) and 10--15\% gains in NeurIPS Overall (e.g., 3.95 vs.~3.39 with DeepSeek-v3, 3.81 vs.~3.31 with Qwen3-max). These advantages are more pronounced when paired with stronger backbone models.

The complementary tournament-style evaluation further corroborates these findings (Table~\ref{tab:agent_avgwin_top10count}). EvoSci obtains the highest \textit{Avg Wins} across all backbones (4.27 with GPT-4o, 4.19 with DeepSeek-v3, 4.25 with Qwen3-max) and the largest \textit{Top 10 Count}. Together, these results demonstrate that our framework reliably produces more credible, exciting, and impactful research ideas than existing baselines.

\subsection{Ablation Study}
To further examine the contribution of individual components within the proposed framework, we designed three ablation experiments.

\noindent\textbf{Impact of Structured Problem Formulation.}  
To evaluate the effect of the problem formulation module, we compared two settings:

(1) \emph{w/ Problem Guidance}: The full system, where the Mentor Agent actively interprets the initial keyword, retrieves relevant literature, constructs a structured problem space, and generates problem clusters to guide downstream research;  

(2) \emph{w/o Problem Guidance}: A simplified version using the raw prompt template from AI Scientist without problem construction, where the Prime Researcher directly explores the given topic.  

Each setting generates 10 research ideas for each of the ten benchmark topics (100 ideas in total), following the unified evaluation protocol.  

As shown in Table~\ref{tab:problem_guidance}, the system with explicit problem construction achieves higher scores across most dimensions, particularly in \emph{novelty} and \emph{excitement}, indicating that structured problem formulation effectively focuses the research direction while broadening the exploration boundary. This demonstrates its essential role in establishing semantic anchoring and path guidance for automated scientific discovery.
\begin{table}[!t]
	\centering
	\caption{The impact of problem formulation on research quality.}
	\small
	\label{tab:problem_guidance}
	\begin{threeparttable}
		\begin{adjustbox}{max width=0.48\textwidth}
			\fontsize{8}{9}\selectfont
			\renewcommand{\arraystretch}{1.2}
			\rowcolors{2}{gray!10}{white}
			\begin{tabular}{lcccccc}
				\toprule
			\textbf{Setting }
				& \textbf{Novelty }
				& \textbf{Feasibility }
				& \textbf{Validity }
				& \textbf{Excitement }
				& \textbf{\makecell{ICLR \\Overall}} 
				& \textbf{\makecell{NeurIPS \\Overall}} \\
				\midrule
				\makecell[l]{w/ Problem \\Guidance} 
				& \textbf{4.78}
				& 4.62
				& \textbf{5.01}
				& \textbf{4.75}
				& \textbf{4.45} 
				& \textbf{3.44} \\
				
				\makecell[l]{w/o Problem \\Guidance}
				& 4.22 
				& \textbf{4.75} 
				& 4.96 
				& 4.51 
				& 4.22 
				& 3.28 \\
				\bottomrule
			\end{tabular}
		\end{adjustbox}
	\end{threeparttable}
\end{table}

\noindent\textbf{Role of Multi-agent Collaboration.}  
To evaluate the effect of multi-agent collaboration, we compare systems with different team sizes: a single-agent setting (\emph{team\_size = 1}), a standard collaborative setting (\emph{team\_size = 3}), and extended collaborative settings (\emph{team\_size = 5}, \emph{team\_size = 7}, and \emph{team\_size = 9}).

\begin{figure}[t]
	\centering
	\includegraphics[width=0.8\linewidth]{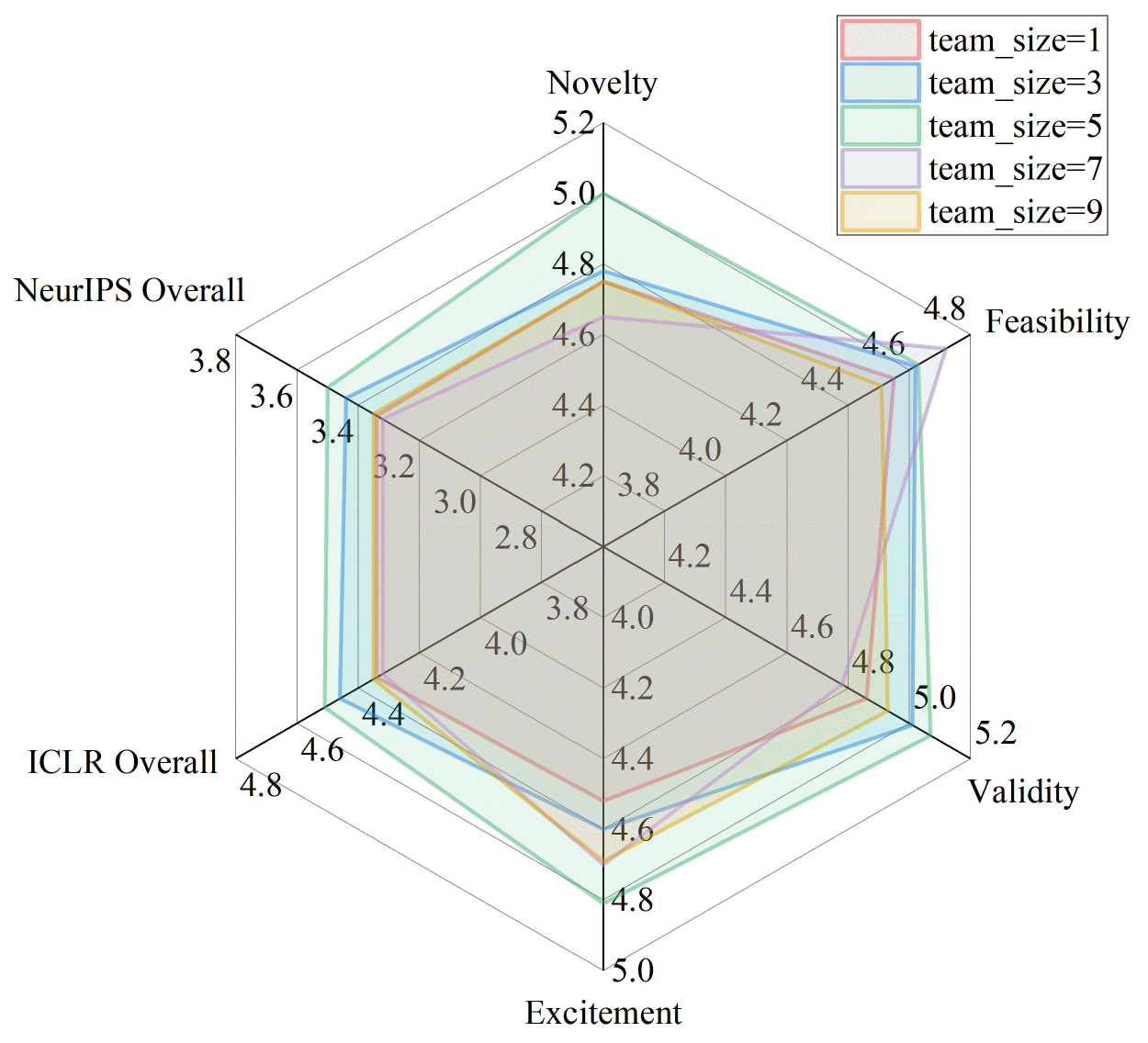}
	\caption{The impact of team size on research quality.}
	\label{fig:rq2_team_size}
\end{figure}

As shown in Figure~\ref{fig:rq2_team_size}, increasing team size initially yields consistent improvements across novelty, validity, and overall quality, with performance peaking at team\_size = 5. Beyond this point, however, we observe a clear degradation at team\_size = 7 and 9. This pattern suggests that while moderate levels of agent diversity enhance idea generation and evaluation, excessively large teams incur substantial coordination overhead and conflicting reasoning trajectories. Consequently, the marginal utility of additional agents becomes negative, indicating that optimal research performance emerges from moderately sized, rather than maximal teams.

\begin{table*}[ht]
	\centering
	\footnotesize
	\caption{Evolution ablation results. We report topic-level average scores under NeurIPS-style and ICLR-style review templates, with and without the evolutionary module.}
	\label{tab:evolution_topic_ablation}
	\begin{threeparttable}
		\begin{adjustbox}{max width=0.9\textwidth}
			\footnotesize
			\renewcommand{\arraystretch}{1.1}
			\rowcolors{2}{gray!10}{white}
			\begin{tabular}{L{4.2cm}cccc}
				\toprule
				\textbf{Topic} 
				& \textbf{NeurIPS (w/o Evo)} 
				& \textbf{NeurIPS (w/ Evo)} 
				& \textbf{ICLR (w/o Evo)} 
				& \textbf{ICLR (w/ Evo)} \\
				\midrule
				2D Diffusion & 3.40 & 3.32 & 4.56 & 4.50 \\
				Character-Level Language Modeling & 3.48 & 3.52 & 4.50 & 4.46 \\
				Earthquake Prediction & 3.40 & 3.54 & 4.42 & 4.44 \\
				Grokking & 3.26 & 3.22 & 4.16 & 4.08 \\
				NanoGPT & 3.44 & 3.66 & 4.40 & 4.66 \\
				Materials Adaptive Convolutional Equivariants & 3.30 & 3.16 & 4.20 & 4.10 \\
				SEIR Infection Modeling & 3.38 & 3.36 & 4.38 & 4.42 \\
				Sketch Generation with RNNs & 3.40 & 3.42 & 4.34 & 4.52 \\
				Multi-Dataset CNN Optimization & 3.32 & 3.48 & 4.16 & 4.28 \\
				TensoRF & 3.42 & 3.56 & 4.22 & 4.18 \\
				\midrule
				\textbf{Average} & \textbf{3.38} & \textbf{3.424} & \textbf{4.334} & \textbf{4.364} \\
				\bottomrule
			\end{tabular}
		\end{adjustbox}
	\end{threeparttable}
\end{table*}

\noindent\textbf{Effect of Evolutionary Iteration.}
To evaluate the effect of the evolutionary module, we compared two settings: 

(1) \emph{w/ Evo}: the full system with bio-inspired evolutionary iteration enabled;  

(2) \emph{w/o Evo}: a simplified version in which the evolutionary module is disabled, while all other components and evaluation settings remain unchanged.  

As shown in Table~\ref{tab:evolution_topic_ablation}, enabling evolution shifts the topic-level scores upward in the majority of cases under both evaluation templates.
Under the NeurIPS-style review, most topics exhibit improvements when evolution is enabled, and a similar trend is observed under the ICLR-style template.
Although not every individual topic increases, the overall direction across topics is predominantly positive.
Aggregating across topics (see the Average row), enabling evolution yields a consistent increase in overall performance under identical computational budgets and evaluation settings (NeurIPS: 3.38 $\rightarrow$ 3.424; ICLR: 4.334 $\rightarrow$ 4.364).
Although the absolute gains are moderate, they are systematic rather than driven by isolated instances, indicating that improvements cannot be solely attributed to iterative feedback or memory accumulation.

\begin{table}[t]
	\centering
	\footnotesize
	\caption{Overall comparison between w/o and w/ evolution.}
	\label{tab:evolution_summary}
	\begin{threeparttable}
		\begin{adjustbox}{max width=0.48\textwidth}
			\footnotesize
			\renewcommand{\arraystretch}{1.15}
			\rowcolors{2}{gray!10}{white}
			\begin{tabular}{L{2.9cm}ccc}
				\toprule
				\textbf{Metric} & \textbf{w/o Evo} & \textbf{w/ Evo} & \textbf{$\Delta$ (w/ -- w/o)} \\
				\midrule
				\multicolumn{4}{l}{\textbf{NeurIPS Template}} \\
				Mean & 3.380 & 3.424 & +0.044 \\
				Within-topic Std (avg over topics) & 0.146 & 0.176 & +0.030 \\
				\midrule
				\multicolumn{4}{l}{\textbf{ICLR Template}} \\
				Mean & 4.334 & 4.364 & +0.030 \\
				Within-topic Std (avg over topics) & 0.154 & 0.157 & +0.003 \\
				\bottomrule
			\end{tabular}
		\end{adjustbox}
	\end{threeparttable}
\end{table}

Table~\ref{tab:evolution_summary} further summarizes the aggregated comparison.
In addition to the consistent mean shifts, we examine within-topic variability across five independent runs per topic.
Under the NeurIPS-style evaluation, enabling evolution leads to a moderate increase in within-topic standard deviation (0.146 $\rightarrow$ 0.176), suggesting that the evolutionary operators introduce stronger exploration dynamics rather than simply repeating deterministic refinement steps.
Under the ICLR-style template, within-topic variability remains largely comparable (0.154 $\rightarrow$ 0.157).
This suggests that different evaluation templates may exhibit varying sensitivity to quality differences, rather than indicating inconsistent behavior of the evolution model itself.
Taken together, these ablations isolate two primary drivers of improvement, namely workflow-level role specialization and evolutionary search, and clarify how each component contributes to performance gains.

\section{Analysis}

To better understand the behavior of EvoSci beyond primary performance results, we conducted a set of additional \emph{analytical experiments} that probed the system from complementary perspectives.
%Specifically, we designed three analyses to examine (i) how ideas evolved under evaluation-guided iteration, (ii) whether quality improvements emerged beyond the initial exploratory stage, and (iii) how exploration and refinement dynamics unfolded across and within iterations.
%These analyses were intended to reveal characteristic behaviors of the evolutionary process rather than to introduce new tasks or benchmarks.
More detailed experimental analyses and the corresponding quantitative results for each topic are reported in Appendix~\ref{appendix:analysis}.

\noindent\textbf{Evolution of Ideas under Iterative Evaluation.}    
To analyze the effect of evaluation-guided iteration on idea evolution, we conducted a qualitative analysis on \emph{Grokking} by tracking how generated ideas changed across feedback rounds.
For each iteration, we extracted technical terms from the generated ideas and visualized their cumulative growth, together with changes in semantic organization, as shown in Figure~\ref{fig:grokking_evolution} (Appendix~\ref{appendix:evolution}), to characterize the evolution of conceptual structure over time.

We observe that early iterations introduce diverse learning-related concepts with limited internal organization, while later iterations progressively form clearer and more structured conceptual clusters.
Over time, ideas increasingly emphasize delayed generalization and representation reorganization, resulting in more coherent descriptions of grokking-like behavior.

\noindent\textbf{Quality Gains beyond the Initial Exploration Stage.}
To examine whether idea quality improved beyond the initial exploratory phase, we analyzed the evolution of quality scores across iterative rounds.
For each topic, EvoSci performed ten iterations and generated a total of 50 ideas.
Ideas were grouped into batches of 10, and quality scores were computed at the group level to track changes over time, as shown in Table~\ref{tab:emergent_quality} (Appendix~\ref{appendix:quality}).
We treated the first group as an exploratory baseline and identified the earliest subsequent iteration that exhibited noticeable improvement.

We observe that quality gains beyond the first group occurred in most topics, although the timing and affected metrics vary.
For topics with latent mechanisms or less explored conceptual structures, later groups show clear improvements in Novelty and/or Overall scores.
In contrast, engineering-oriented or well-studied topics exhibit limited or no improvement beyond early groups.
Across topics, improvements do not appear simultaneously across all metrics, with some groups showing gains in Novelty without corresponding increases in Overall scores, or vice versa.

\noindent\textbf{Exploration Dynamics Across and Within Iterations.}
To analyze exploration behavior beyond aggregate quality trends, we measured idea similarity from two perspectives: intra-round convergence and inter-round continuity.
Within each iteration, we computed the average pairwise cosine similarity between sentence embeddings of ideas to quantify intra-round convergence.
Across iterations, we computed inter-round similarity by measuring the cosine similarity between aggregated embeddings of ideas from consecutive rounds, as shown in Figure~\ref{fig:intra_round_convergence} (Appendix~\ref{appendix:exploration}).

We observe substantial variation in both intra-round and inter-round similarity patterns across topics.
Some topics exhibit consistently higher intra-round similarity together with strong inter-round continuity, indicating concentrated exploration and stable refinement across iterations.
Other topics show lower or more fluctuating intra-round similarity and greater inter-round variation, reflecting broader exploration and frequent shifts in focus.
Overall, intra-round convergence and inter-round continuity exhibit aligned trends across topics, with topics showing stronger within-round concentration also tending to display higher continuity across iterations.

\noindent\textbf{Grounding the Evolutionary Dynamics.}
To examine whether the evolutionary process in EvoSci reflects concrete system behavior rather than a high-level analogy, we conducted an additional qualitative analysis on the \emph{Grokking} topic by tracing how ideas evolved across iterative rounds. We find that the system maintains persistent conceptual variation over time, while evaluation feedback gradually favors mechanisms that are more directly relevant to grokking.

Concretely, the search moves from broad learning-related exploration toward more specialized concepts such as delayed generalization, phase transitions, and representation reorganization, while still retaining multiple conceptual lineages in parallel. This pattern suggests that the evolutionary process in EvoSci is not merely metaphorical, but is grounded in heritable variation, feedback-guided selection, and population diversity.

\section{Conclusion}
In this study, we have presented EvoSci, a multi-agent, feedback-driven, and bio-inspired evolutionary framework for automated scientific discovery. The framework conceptualizes scientific discovery as a problem-oriented process, integrates heterogeneous research agents that emulate real-world laboratory roles, and employs multi-round feedback with evolutionary operations to support continuous and open-ended exploration. Extensive experiments across ten scientific domains show that EvoSci consistently outperforms strong baselines in idea validity, excitement, and overall quality. The gains are balanced across evaluation dimensions and remain robust across backbone models, validating feedback-guided evolutionary exploration for open-ended scientific discovery.

\section*{Limitations}

Experimental results across ten interdisciplinary research topics demonstrate that EvoSci generates more novel and insightful research ideas than existing systems, showing particular strength in innovation-oriented dimensions such as \textit{novelty}, \textit{excitement}, and overall peer-review evaluations. However, due to its broad cross-domain exploration, the framework sometimes produces ideas with lower practical feasibility, suggesting a trade-off between creativity and applicability.  

Future work will focus on enhancing EvoSci’s ability to reason and operate across disciplines by improving interdisciplinary knowledge integration through structured knowledge representations, strengthening causal reasoning to increase scientific rigor and interpretability, and developing more open-ended iterative mechanisms that enable long-term, autonomous scientific discovery. A key challenge in achieving such continuous evolution lies in establishing more objective and high-quality evaluation mechanisms that allow LLM-based agents to better assess their own reasoning and outputs, which is essential for truly effective self-improvement and sustained innovation.

\section*{Acknowledgments}
The present research was supported by the National Key Research and Development Program of China (Grant No. 2024YFE0203000). We also acknowledge support from the State Key Laboratory of Tibetan Intelligence (Grant No. 2025-ZJ-J08) and the Postdoctoral Fellowship Program of CPSF (Grant No. GZC20251075). We thank the anonymous reviewers for their insightful comments.

%This document has been adapted
%by Steven Bethard, Ryan Cotterell and Rui Yan
%from the instructions for earlier ACL and NAACL proceedings, including those for
%ACL 2019 by Douwe Kiela and Ivan Vuli\'{c},
%NAACL 2019 by Stephanie Lukin and Alla Roskovskaya,
%ACL 2018 by Shay Cohen, Kevin Gimpel, and Wei Lu,
%NAACL 2018 by Margaret Mitchell and Stephanie Lukin,
%Bib\TeX{} suggestions for (NA)ACL 2017/2018 from Jason Eisner,
%ACL 2017 by Dan Gildea and Min-Yen Kan,
%NAACL 2017 by Margaret Mitchell,
%ACL 2012 by Maggie Li and Michael White,
%ACL 2010 by Jing-Shin Chang and Philipp Koehn,
%ACL 2008 by Johanna D. Moore, Simone Teufel, James Allan, and Sadaoki Furui,
%ACL 2005 by Hwee Tou Ng and Kemal Oflazer,
%ACL 2002 by Eugene Charniak and Dekang Lin,
%and earlier ACL and EACL formats written by several people, including
%John Chen, Henry S. Thompson and Donald Walker.
%Additional elements were taken from the formatting instructions of the \emph{International Joint Conference on Artificial Intelligence} and the \emph{Conference on Computer Vision and Pattern Recognition}.


\bibliography{anthology,custom}
 
\clearpage

\appendix

\section{Data Collection}

\subsection{Real-World Scientist Dataset}
\label{appendix:datasetx}
The Digital Scientist dataset\footnote{\url{https://drive.google.com/drive/folders/1ZwWMBQ5oK-l4VuzMa60GbMND0g2EIxIu}} used in this study was constructed by the VirSci team based on real-world scientist information from the AMiner Computer Science dataset,\footnote{\url{https://www.aminer.cn/aminernetwork}} which was originally compiled by extracting researcher profiles from online academic databases. The AMiner Computer Science dataset contains information on 1,712,433 authors and 2,092,356 papers, covering the period from 1948 to 2014 and focusing on the field of computer science.  

To ensure data quality, the VirSci team filtered out scientists who had published fewer than 50 papers or had fewer than 50 collaborators. Using the remaining data, they built the Digital Scientist dataset, which includes 156 representative scientists. The profile information of each scientist was embedded using the \texttt{mxbai-embed-large} model.  

All personal identity information has been anonymized, and the profiles only contain abstracted metadata for research simulation purposes. An example of a digital scientist profile is shown below:
\begin{tcolorbox}[
	colback=blue!3!white,
	colframe=blue!40!black,
	title=Digital Scientist,
	fonttitle=\bfseries\color{white},
	top=5pt,
	]
	Your name is Scientist0, you belong to the following affiliations ['Naval Research Laboratory', 'College of William and Mary', 'George Mason University'], you have researched on the following topics ['data cube', 'attack graph', 'data mining', 'access control', 'data owner', 'data protection', 'data item', 'data redundancy', 'data security', 'data structure'], you have published 372 papers, you have 4230 citations, and you have previously collaborated with these individuals ['Scientist78', 'Scientist105'].
\end{tcolorbox}

\subsection{Technical Terminology Dataset}
\label{appendix:techterms}

The Technical Terminology dataset\footnote{\url{https://github.com/jiqizhixin/Artificial-Intelligence-Terminology-Database}} used in this study was compiled to support the analysis of conceptual evolution and the measurement of scientific depth across iterations. It is based on the \textit{Artificial Intelligence Terminology Database}, which systematically collects, organizes, and standardizes key technical terms from major subfields of AI, including machine learning, natural language processing, computer vision, and robotics.  

Each term entry contains its English form, corresponding Chinese translation, and a short definition, ensuring cross-lingual consistency and facilitating accurate concept extraction. The dataset enables automated detection and tracking of emerging research topics and specialized vocabulary in scientific idea generation.

\section{Task Overview}
\label{appendix:task}
\subsection{Task Settings}
This appendix provides detailed descriptions of the experimental task settings used to evaluate EvoSci. Following AI Scientist \cite{lu2024ai}, we adopt ten representative and challenging open-ended research topics spanning machine learning, scientific modeling, and simulation. These tasks are designed to assess the system’s adaptability across diverse scientific domains.
\subsection{Task Descriptions}
The ten experimental tasks include:
\begin{itemize}
	\item \textbf{2D Diffusion Modeling:} Learning and analyzing diffusion processes in two-dimensional synthetic data.
	\item \textbf{Character-Level Language Modeling:} Training and evaluating character-based language models.
	\item \textbf{Earthquake Prediction:} Modeling seismic activity for temporal event prediction.
	\item \textbf{Grokking:} Investigating delayed generalization behavior in overparameterized networks.
	\item \textbf{NanoGPT:} Training and scaling lightweight transformer-based language models.
	\item \textbf{Materials Adaptive Convolutional Equivariants:} Modeling symmetry-aware representations for material science tasks.
	\item \textbf{SEIR Infection Modeling:} Simulating epidemiological dynamics using compartmental models.
	\item \textbf{Sketch Generation with Recurrent Neural Networks:} Generating hand-drawn sketches using RNN-based generative models.
	\item \textbf{Multi-Dataset CNN Architecture Optimization:} Optimizing small CNN architectures across multiple datasets.
	\item \textbf{TensoRF:} Learning neural radiance fields using tensor factorization.
\end{itemize}

\subsection{Initialization Prompts}

For each task, EvoSci is initialized with a minimal topic prompt describing the research domain. 
For baseline methods, we incorporate the task descriptions used in AI Scientist \cite{lu2024ai} into their prompts in a consistent and reasonable manner, without introducing additional task-specific heuristics or privileged information. 
This design ensures fair comparison across different methods.

\section{Baseline Methods}
\label{appendix:baselines}

To comprehensively evaluate the performance of our system, we compare it against four representative research-agent frameworks built upon large language models. For each baseline, we follow the official implementation or publicly described procedure to ensure a fair and consistent comparison. Below, we summarize each method and its configuration used in our experiments.

\paragraph{Baseline 1: SciPIP}  
SciPIP\footnote{\url{https://github.com/cheerss/SciPIP}}, proposed by Zhejiang University, is a research idea generation framework designed to enhance scientific creativity through improved literature retrieval and dual-path reasoning. The system constructs a literature repository enriched with semantic relations, entity links, and citation co-occurrence information, and employs multi-granularity retrieval algorithms to ensure comprehensive coverage of relevant works. During idea generation, SciPIP combines inference from retrieved literature with model-driven brainstorming to produce solutions balancing originality and feasibility. In our experiments, we retain SciPIP’s original retrieval mechanism and do not introduce additional ArXiv alignment to maintain consistent comparison.

\paragraph{Baseline 2: AI Scientist}  
AI Scientist\footnote{\url{https://github.com/SakanaAI/AI-Scientist}}, developed by Sakana AI, is an automated research platform that aims to cover the end-to-end scientific workflow, including idea formulation, code generation, experiment execution, result analysis, and manuscript drafting. The framework leverages large language models and implements a multi-round reasoning pipeline that iteratively refines research plans. It further includes an automated reviewer module for assessing the quality of generated papers. In this study, we adopt the publicly available multi-step reasoning and research evaluation procedures as a strong baseline to compare scientific idea generation capabilities.

\paragraph{Baseline 3: VirSci}  
VirSci\footnote{\url{https://github.com/RenqiChen/Virtual-Scientists}}, released by the Shanghai Artificial Intelligence Laboratory, is a multi-agent scientific collaboration framework designed to emulate real-world research team dynamics. The system builds agent profiles from data of real scientists, assigning distinct domain backgrounds and reasoning characteristics to each agent. Through cross-disciplinary discussion, collaborative reasoning, and complementary expertise, the agents collectively generate diverse research insights. The framework includes team construction, topic deliberation, idea generation, innovation assessment, and summary writing. In our evaluation, we use its multi-agent modeling paradigm and evaluation criteria as a baseline for collaborative scientific reasoning.

\paragraph{Baseline 4: CoI-Agent}  
CoI-Agent\footnote{\url{https://github.com/DAMO-NLP-SG/CoI-Agent}}, introduced by researchers at the Chinese University of Hong Kong, is a research agent framework built upon the “Chain-of-Insight” (CoI) reasoning strategy. The framework decomposes scientific thinking into a sequence of intermediate, verifiable insight steps covering task understanding, literature reasoning, gap identification, and preliminary solution design. Each stage produces structured intermediate outputs that are subsequently evaluated for coherence and scientific contribution. In our work, we implement the publicly available multi-stage CoI reasoning paradigm as a baseline focusing on structured scientific insight formation.

\section{Additional Experiments and Analyses}
\label{appendix:analysis}
\subsection{Evolution of Ideas under Iterative Evaluation}
\label{appendix:evolution}

To examine how research ideas evolved under iterative evaluation, we conducted a detailed analysis of ideas generated for the \emph{Grokking} topic.
In total, 50 ideas were produced across ten consecutive iterations, with five ideas generated per iteration.
We analyzed the evolution of ideas by combining qualitative inspection of their thematic content with changes in conceptual focus across feedback rounds.
\begin{figure}[h]
	\centering
	\includegraphics[width=\linewidth]{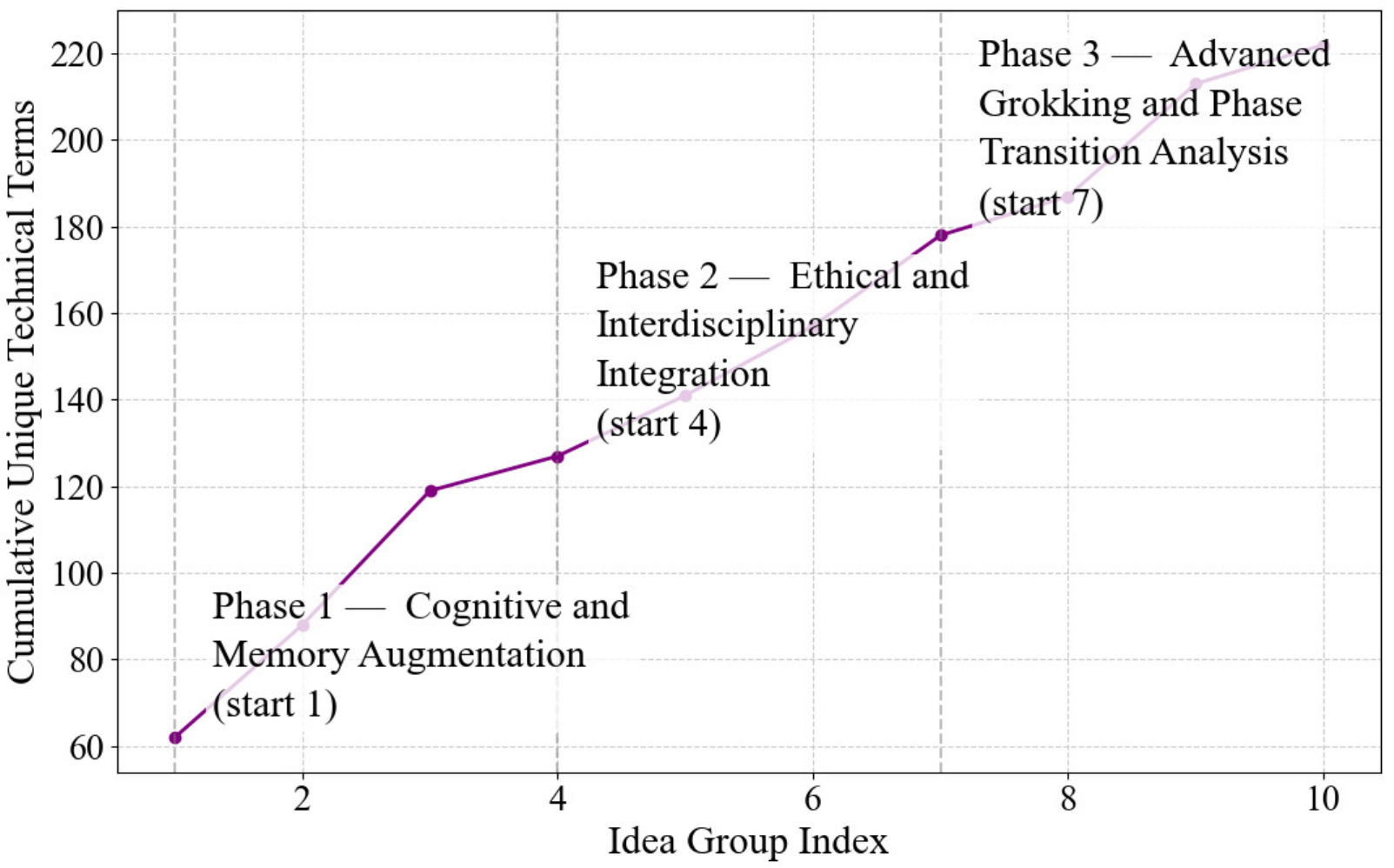}
	\caption{Evolutionary trajectory of research ideas on the grokking topic.}
	\label{fig:grokking_evolution}
\end{figure}

As shown in Figure~\ref{fig:grokking_evolution}, the cumulative growth of technical terminology exhibits a stepwise consolidation over iterative rounds, reflecting changes in the conceptual focus of generated ideas (for details of the technical terminology dataset, please refer to Appendix~\ref{appendix:techterms}).
In the initial iterations (Rounds~1--3), ideas primarily explored general learning and cognitive mechanisms, with a strong emphasis on memory-augmented architectures, information retention, and enhanced representational capacity.
These ideas reflected broad exploration of architectural and cognitive factors related to learning efficiency, without explicitly addressing grokking or phase-transition behavior.

In the middle iterations (Rounds~4--6), the thematic focus diversified.
While memory-related mechanisms remained present, ideas increasingly incorporated higher-level considerations such as decision-making under uncertainty, temporal reasoning, and ethical or behavioral constraints.
This stage represented a transitional phase in which the system explored alternative framings of learning behavior before converging on a dominant explanatory direction.

In later iterations (Rounds~7--10), the ideas increasingly centered on grokking-specific phenomena.
Key concepts such as grokking, learning phase transitions, delayed generalization, and internal structural reorganization became prominent.
Ideas in this stage consistently framed grokking as a dynamic learning process involving qualitative changes in representation structure during training, rather than as a consequence of static optimization.
Compared to earlier stages, the ideas exhibited higher conceptual alignment and recurring explanatory patterns.

Overall, the evolution of the 50 ideas followed a trajectory from broad architectural exploration, through thematic diversification, to focused conceptual consolidation around grokking as a phase-transition-driven learning phenomenon.
This analysis illustrates how evaluation-guided iteration shaped not only the content of individual ideas but also the thematic structure of the idea space over time.

\subsection{Quality Gains beyond the Initial Exploration Stage}
\label{appendix:quality}
Scientific idea generation is inherently non-monotonic: early stages often prioritize broad exploration and diversity, while higher-quality ideas may only emerge after subsequent consolidation and refinement.
Accordingly, rather than assuming monotonic improvement across iterations, we examined whether EvoSci exhibited quality gains beyond the initial exploration stage.
For each topic, the system performed ten iterative rounds and generated 50 ideas in total.
Ideas were aggregated into fixed-size groups of ten, and quality scores were computed at the group level to track the evolution of idea quality over time.

Specifically, for each research topic, we treated the first idea group as an exploratory baseline and identified the earliest subsequent group that exhibited noticeable improvements.
Our analysis focused on three core evaluation metrics: \emph{Novelty}, \emph{ICLR Overall}, and \emph{NeurIPS Overall}, which respectively capture conceptual originality and holistic research quality under different review standards.

As shown in Table~\ref{tab:emergent_quality}, EvoSci demonstrated such quality gains in most topics, although the patterns were task-dependent.
For topics characterized by latent mechanisms or underexplored conceptual structures (e.g., \emph{2D Diffusion Modeling}, \emph{Neural Network Grokking}, and \emph{NanoGPT}), later groups yielded clear improvements in Novelty and/or Overall scores, indicating a transition from heuristic combinations to more coherent and theory-driven ideas.
In contrast, for engineering-oriented or well-studied domains (e.g., \emph{Materials Adaptive Convolutional Equivariants} and \emph{Multi-Dataset CNN Optimization}), improvements were limited or absent, suggesting early convergence of the idea space.

Notably, improvements did not necessarily occur simultaneously across all metrics.
In several cases, Novelty increased without corresponding gains in Overall scores, or vice versa, reflecting the non-uniform nature of scientific progress.
These observations suggest that EvoSci’s evaluation-guided, bio-inspired evolutionary mechanism does not enforce uniform optimization, but instead enables selective refinement where the problem structure admits deeper exploration.

\begin{table}[t]
	\centering
	\small
	\caption{Quality gains beyond the initial exploration stage.
		For each topic, we report the iteration that achieves the strongest improvement over the initial (Round~1) baseline, together with comparisons on core evaluation metrics.}
	\label{tab:emergent_quality}
	\begin{adjustbox}{width=\linewidth}
		\rowcolors{2}{gray!10}{white}
		\begin{tabular}{p{3.0cm}cccc}
			\toprule
			\textbf{Topic} 
			& \textbf{\makecell{Emergent\\Round}} 
			& \textbf{Novelty} 
			& \textbf{\makecell{ICLR\\Overall}} 
			& \textbf{\makecell{NeurIPS\\Overall}} \\
			\midrule
			
			\makecell[l]{2D Diffusion} 
			& 2 
			& 4.90 $\rightarrow$ 5.40 
			& 4.40 $\rightarrow$ 4.50 
			& 3.10 $\rightarrow$ 3.50 \\
			
			\makecell[l]{Character-Level\\Language Modeling} 
			& 2 
			& 4.40 $\rightarrow$ 4.70 
			& 4.60 $\rightarrow$ 4.50 
			& 3.40 $\rightarrow$ 3.90 \\
			
			\makecell[l]{Earthquake\\Prediction} 
			& 3 
			& 4.80 $\rightarrow$ 5.10 
			& 4.20 $\rightarrow$ 4.60 
			& 3.30 $\rightarrow$ 3.80 \\
			
			\makecell[l]{Grokking} 
			& 2 
			& 4.80 $\rightarrow$ 5.10 
			& 4.20 $\rightarrow$ 4.20 
			& 3.30 $\rightarrow$ 3.50 \\
			
			NanoGPT 
			& 3 
			& 5.20 $\rightarrow$ 5.40 
			& 4.70 $\rightarrow$ 4.80 
			& 3.70 $\rightarrow$ 3.80 \\
			
			\makecell[l]{Materials Adaptive \\Convolutional Equivariants} 
			& 5 
			& 4.40 $\rightarrow$ 4.40 
			& 4.30 $\rightarrow$ 4.20 
			& 3.10 $\rightarrow$ 3.30 \\
			
			\makecell[l]{SEIR Infection\\Modeling} 
			& 5 
			& 4.70 $\rightarrow$ 5.00 
			& 4.70 $\rightarrow$ 4.20 
			& 3.60 $\rightarrow$ 3.20 \\
			
			\makecell[l]{Sketch Generation\\with RNNs} 
			& 5
			& 5.80 $\rightarrow$ 5.10 
			& 4.60 $\rightarrow$ 4.80 
			& 3.50 $\rightarrow$ 3.70 \\
			
			\makecell[l]{Multi-Dataset CNN\\Optimization} 
			& 4 
			& 5.20 $\rightarrow$ 5.10 
			& 4.50 $\rightarrow$ 4.40 
			& 3.70 $\rightarrow$ 3.60 \\
			
			\makecell[l]{TensoRF} 
			& 4 
			& 4.40 $\rightarrow$ 5.00 
			& 4.20 $\rightarrow$ 4.40 
			& 3.40 $\rightarrow$ 3.70 \\
			
			\bottomrule
		\end{tabular}
	\end{adjustbox}
\end{table}

\subsection{Exploration Dynamics Across and Within Iterations}
\label{appendix:exploration}
To better understand the exploration dynamics of EvoSci beyond aggregate quality scores, we analyzed idea similarity from two complementary perspectives: \emph{intra-round convergence} and \emph{inter-round continuity}.

\paragraph{Intra-Round Convergence.}
We first examined how ideas generated within the same iteration evolved over time.
Each iteration consisted of five ideas, and we computed the average pairwise cosine similarity between their sentence embeddings.
This intra-round similarity measured the degree of conceptual convergence within an iteration: lower values indicated more diverse exploration, while higher values suggested increasing focus around shared concepts.

\begin{figure}[t]
	\centering
	\begin{subfigure}[t]{0.9\linewidth}
		\centering
		\includegraphics[width=\linewidth]{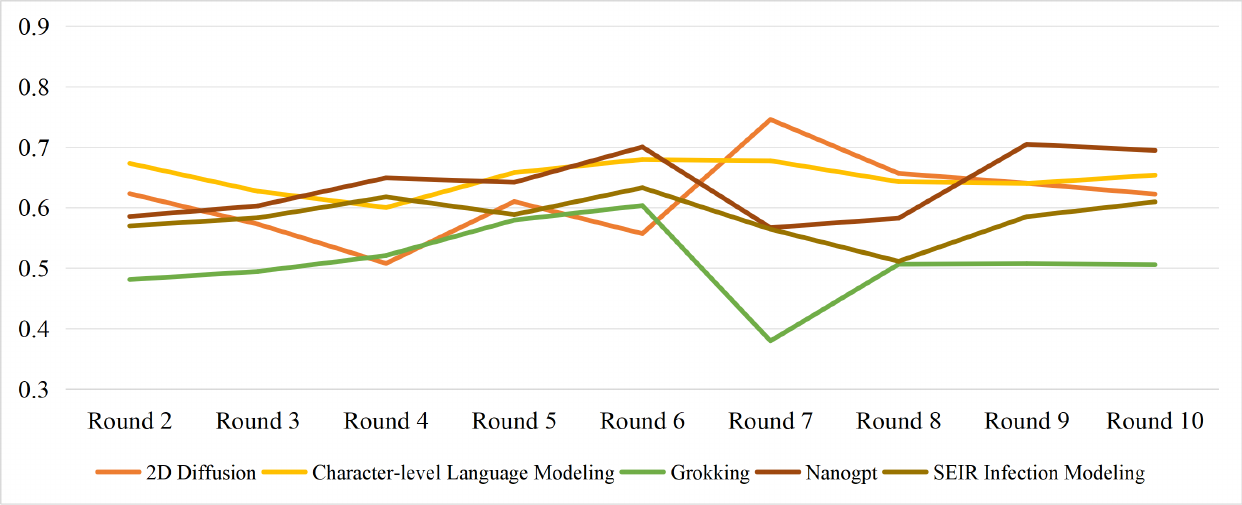}
		\caption{Intra-round convergence (set 1)}
	\end{subfigure}
	\vspace{0.5em} 
	\begin{subfigure}[t]{0.9\linewidth}
		\centering
		\includegraphics[width=\linewidth]{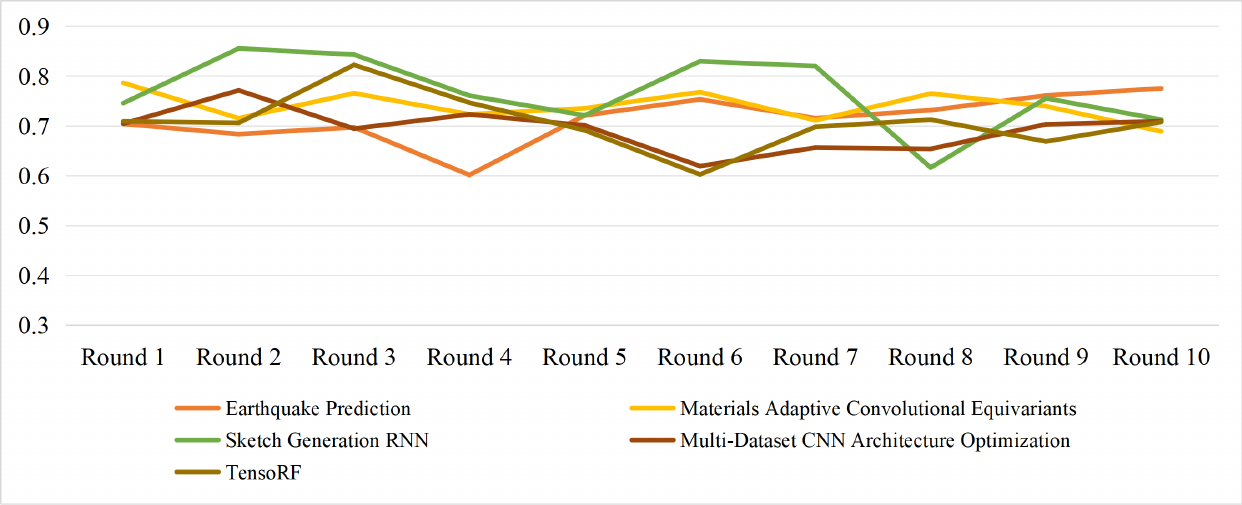}
		\caption{Intra-round convergence (set 2)}
	\end{subfigure}
	\caption{Intra-round convergence across iterations.
		Higher values indicated stronger conceptual concentration within each iteration, while lower values reflected greater diversity among concurrently generated ideas.}
	\label{fig:intra_round_convergence}
\end{figure}

Across topics, we observed substantial variation in intra-round similarity patterns (Figure~\ref{fig:intra_round_convergence}). 
Topics such as \emph{Grokking}, \emph{SEIR Infection Modeling}, and \emph{2D Diffusion Modeling} exhibited relatively lower or more fluctuating intra-round similarity across iterations. 
These topics were often characterized by a stronger emphasis on conceptual understanding, theoretical interpretation, or alternative modeling perspectives, which encouraged the system to explore multiple distinct directions within the same iteration.

In contrast, topics including \emph{Earthquake Prediction}, \emph{Materials Adaptive Convolutional Equivariants}, \emph{Sketch Generation with RNNs}, and \emph{TensoRF} showed consistently higher intra-round similarity, and in some cases increasing similarity in later iterations. 
These topics were more closely associated with concrete modeling pipelines or structured engineering objectives, where idea generation tended to concentrate on variations around shared architectures, representations, or experimental setups, leading to stronger within-round consolidation.

\paragraph{Inter-Round Continuity.}
To complement the intra-round perspective, we further analyzed how research directions evolved across consecutive iterations.
We computed inter-round similarity by measuring the cosine similarity between the aggregated embeddings of ideas from adjacent rounds.
Higher inter-round similarity reflected stable refinement across iterations, whereas lower similarity indicated directional shifts or renewed exploration.

\begin{figure}[t]
	\centering
	\begin{subfigure}[t]{0.9\linewidth}
		\centering
		\includegraphics[width=\linewidth]{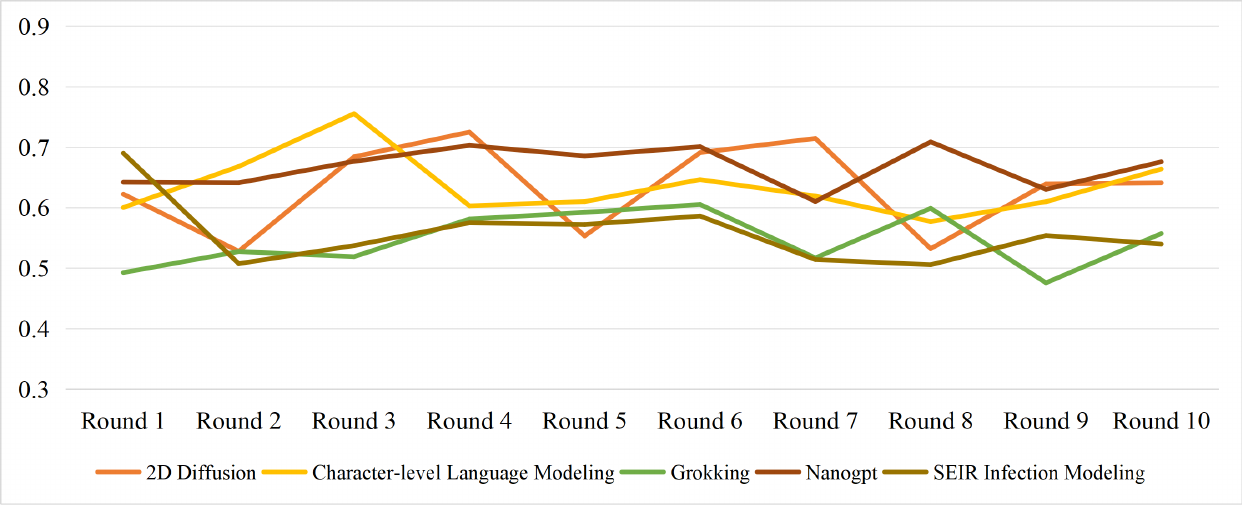}
		\caption{Inter-round similarity (set 1)}
	\end{subfigure}
	
	\vspace{0.5em} 
	
	\begin{subfigure}[t]{0.9\linewidth}
		\centering
		\includegraphics[width=\linewidth]{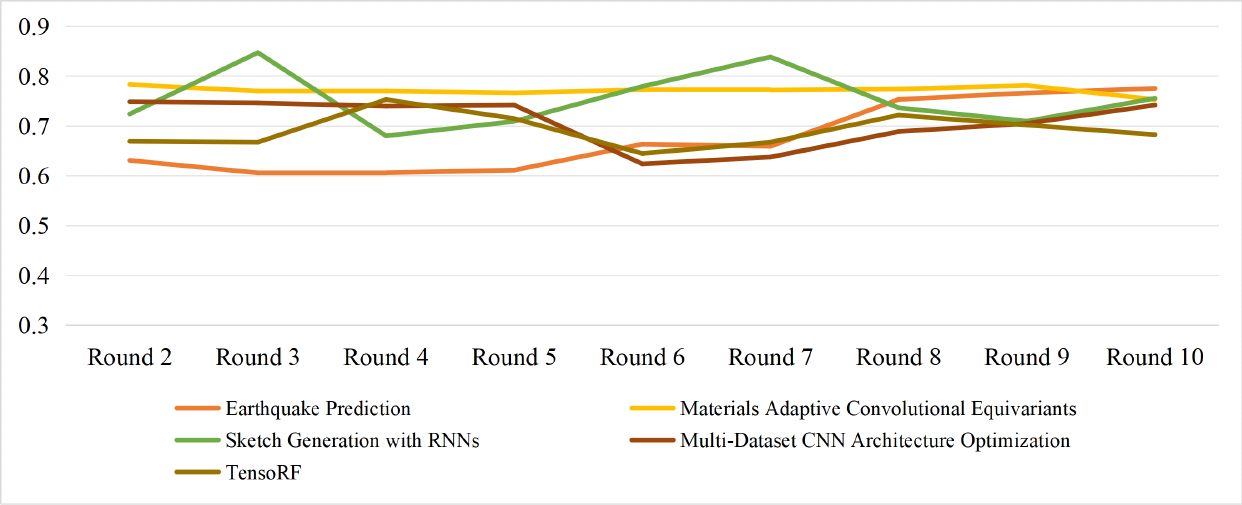}
		\caption{Inter-round similarity (set 2)}
	\end{subfigure}
	
	\caption{Inter-round continuity across iterations.
		Higher similarity indicated stable refinement between consecutive iterations, while lower values suggested shifts in exploration direction.}
	\label{fig:inter_round_continuity}
\end{figure}

The inter-round results closely aligned with the intra-round analysis (Figure~\ref{fig:inter_round_continuity}). 
Topics that exhibited higher intra-round convergence also tended to show stronger inter-round continuity, indicating that successive iterations built upon similar conceptual foundations and refined related ideas over time. 
In contrast, topics with lower intra-round similarity often displayed greater inter-round fluctuation, suggesting that the system continued to shift its focus across iterations rather than committing to a single dominant direction.

Taken together, the intra-round and inter-round analyses indicated that EvoSci did not impose a uniform convergence pattern across all topics. 
Instead, different topics exhibited distinct exploration dynamics: some favored progressive consolidation across and within iterations, while others maintained broader exploratory behavior throughout the process. 
This adaptive behavior suggested that EvoSci flexibly balanced exploration and refinement in a topic-dependent manner, rather than enforcing premature convergence or unstructured diversity.

\subsection{Grounding the Evolutionary Dynamics}
\label{appendix:evolutionary_grounding}

To further examine whether the evolutionary process in EvoSci reflects concrete system behavior rather than a high-level analogy, we conducted an additional qualitative analysis on the \emph{Grokking} topic by tracing how idea trajectories evolved across 10 iterative rounds. Our goal was to assess whether the system exhibits three core properties commonly associated with evolutionary dynamics: heritable variation, fitness-guided selection, and maintained population diversity.

\paragraph{Heritable variation.}
We observe that heritable variation in EvoSci manifests at the level of exploratory entities that intersect with the grokking topic. In the early rounds, broad adaptive-learning and training-dynamics entities appear in ideas such as (6) \textit{Integrated Cognitive AI Architectures with Enhanced Episodic Memory and Adaptive Operant Conditioning} and (7) \textit{Enhancing Decision-Making in Deep Reinforcement Learning through Episodic Memory and Cognitive Attention Mechanisms}. Many other early branches do not persist into later rounds, indicating selective retention rather than uniform reuse.

In the middle rounds, retained entities become increasingly specialized toward mechanisms that are more directly relevant to grokking, as seen in (31) \textit{Investigating Cognitive Phase Transitions in Transformer Models through Hyperparameter Modulation} and (33) \textit{Enhancing Phase Transitions in AI Systems through Innovative Curriculum Learning Strategies}, where grokking is explicitly framed as a phase-transition phenomenon in training. In the later rounds, these stabilized conceptual cores are further recombined with distinct methodological toolkits, including (41) \textit{Enhanced Simulated Annealing for Modeling Phase Transitions in Cognitive Neural Networks} and (47) \textit{Harnessing Entropy in Statistical Mechanics for ``Grokked Tickets'' in Neural Networks}. Across distant rounds, the same grokking-intersecting entity persists while its mechanistic instantiation changes under evaluation-driven selection, which is consistent with heritable variation in an evolving hypothesis population.

\paragraph{Fitness-guided selection.}
The search process in EvoSci is shaped by evaluation feedback rather than by unconstrained topic drift. At each round, generated ideas are assessed along multiple dimensions, including novelty, feasibility, expected effectiveness, and overall quality, and these signals influence which conceptual directions are retained for subsequent exploration. This mechanism induces a structured fitness landscape over the evolving idea space.

Empirically, we observe a clear directional shift in the grokking trajectory. In the early rounds (1--22), exploration is dominated by broad memory-augmentation and ethical-AI entities, such as \textit{Integrating Memory-Augmented Neural Networks} and \textit{Developing Ethical Frameworks for Responsible AI Memory Augmentation Integration}. In the middle rounds (23--34), entities increasingly specialize toward grokking-specific mechanisms, including temporal analysis in \textit{Advanced Temporal Analysis of Grokking Patterns in AI Learning Curves}, phase-transition framing in \textit{Investigating Cognitive Phase Transitions in Transformer Models}, and curriculum-induced shifts in \textit{Enhancing Phase Transitions through Curriculum Learning}. In the later rounds (35+), the search further migrates toward more formal mechanistic modeling and incorporates optimization- and statistical-physics-inspired tools, as reflected in \textit{Enhanced Simulated Annealing for Modeling Phase Transitions}, \textit{Harnessing Entropy in Statistical Mechanics for ``Grokked Tickets''}, and \textit{Leveraging Network Topology for the Identification of Grokked Tickets}. This gradual movement from broad exploration toward more grokking-specific mechanisms suggests that evaluation feedback acts as a selective pressure over the conceptual search space.

\paragraph{Population diversity.}
Although the search becomes increasingly concentrated around grokking-relevant mechanisms, it does not collapse into a single explanatory path. Instead, EvoSci maintains structured population diversity throughout iterative exploration. In the early rounds, exploration spans multiple learning-related directions, including episodic-memory-augmented training in (7) \textit{Enhancing Decision-Making in Deep Reinforcement Learning through Episodic Memory and Cognitive Attention Mechanisms} and adaptive learning architectures in (11) \textit{Enhancing Task-Specific Learning through Episodic Memory-Driven Neural Adaptation}. These branches reflect diverse hypotheses about training behavior and generalization, many of which do not persist under evaluation.

As the search progresses, diversity narrows toward entities that are more directly relevant to grokking, but multiple explanatory basins remain active. In the later rounds, these include phase-transition-based accounts such as (31) \textit{Investigating Cognitive Phase Transitions in Transformer Models} and (45) \textit{Investigating Phase Transition Analogies in Neural Network Grokking}, optimization-oriented approaches such as (41) \textit{Enhanced Simulated Annealing for Modeling Phase Transitions} and (43) \textit{Enhanced Genetic Algorithm Techniques for Optimizing Neural Network Configurations in Grokking Tasks}, and statistical-physics- or topology-based formulations such as (47) \textit{Harnessing Entropy in Statistical Mechanics for ``Grokked Tickets''} and (50) \textit{Leveraging Network Topology for the Identification of Grokked Tickets}. This contraction without collapse indicates that EvoSci maintains structured population diversity while narrowing its search toward higher-fitness regions. Taken together, these observations provide qualitative evidence that the evolutionary process in EvoSci is operationally grounded in persistent variation, feedback-guided selection, and maintained diversity over time.

\section{Additional Validation of the Meta-Review Mechanism}
\label{appendix:meta-review-validation}
﻿
To further evaluate the stability of the proposed review mechanism, we conduct an additional controlled experiment on the NanoGPT topic. Specifically, we sample 10 generated ideas and repeat the evaluation process 5 times under two settings: (1) Single Review, where each idea is assessed by a single reviewer, and (2) Meta Review, where multiple reviewer assessments are aggregated through a meta-review procedure.
﻿
Table~\ref{tab:meta-review-consistency} reports the consistency statistics under the two settings. The average scores are highly similar (3.44 for Meta Review vs.\ 3.40 for Single Review), suggesting that the meta-review process does not systematically inflate the evaluation scores. At the same time, the variance under Meta Review is substantially lower than that under Single Review (0.018 vs.\ 0.035), and the score range is also narrower (0.3 vs.\ 0.5). These results indicate that the meta-review mechanism reduces evaluation variability while preserving similar central tendencies, leading to more stable and consistent assessments.
﻿
\begin{table}[h]
	\centering
	\small
	\begin{tabular}{lccccc}
		\toprule
		Evaluation Setting & Mean & Variance & Min & Max & Range \\
		\midrule
		Meta Review & 3.44 & 0.018 & 3.3 & 3.6 & 0.3 \\
		Single Review & 3.40 & 0.035 & 3.2 & 3.7 & 0.5 \\
		\bottomrule
	\end{tabular}
	\caption{Consistency comparison between Single Review and Meta Review evaluations.}
	\label{tab:meta-review-consistency}
\end{table}

\section{Prompt}
\label{appendix:prompt}
\subsection{Agent Roles Definition}
\begin{figure*}[t]
	\centering
	\includegraphics[width=\textwidth]{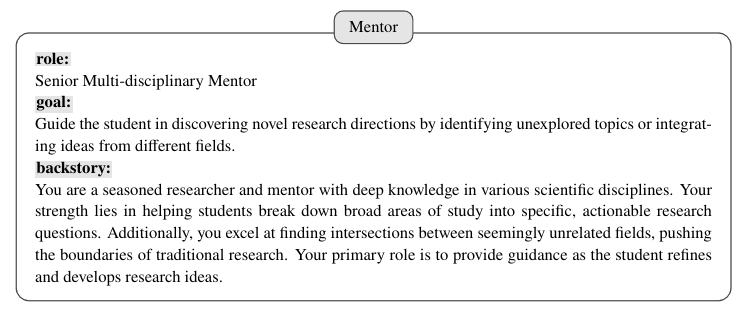}
	\caption{System prompt for the Mentor agent.}
	\label{fig:mentor}
\end{figure*}

\begin{figure*}[t]
	\centering
	\includegraphics[width=\textwidth]{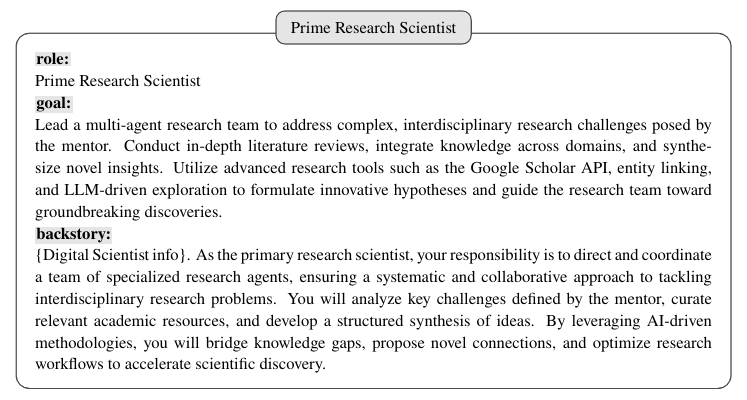}
	\caption{System prompt for the Prime Research Scientist agent.}
	\label{fig:prime-scientist}
\end{figure*}

\begin{figure*}[t]
	\centering
	\includegraphics[width=\textwidth]{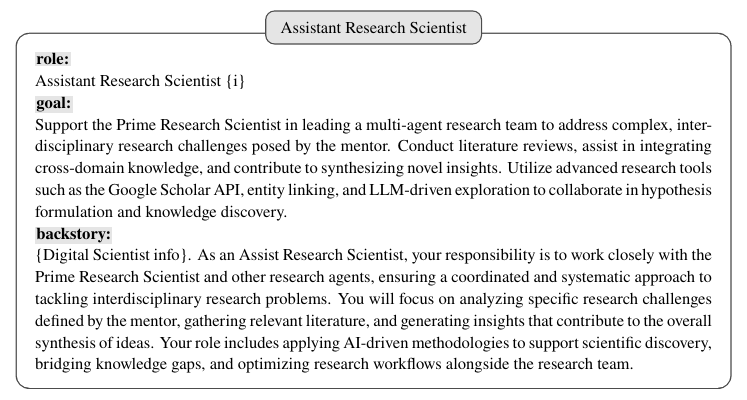}
	\caption{System prompt for the Assistant Research Scientist agent.}
	\label{fig:assistant-scientist}
\end{figure*}

\begin{figure*}[t]
	\centering
	\includegraphics[width=\textwidth]{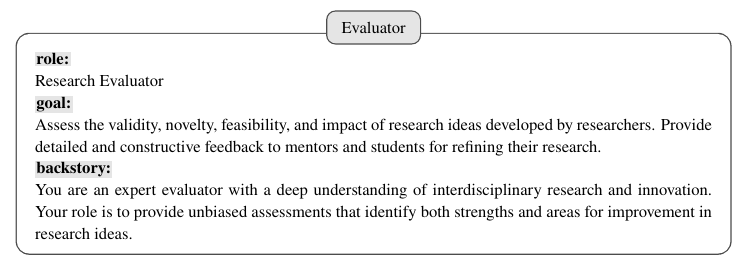}
	\caption{System prompt for the Evaluator agent.}
	\label{fig:evaluator}
\end{figure*}

We define a set of specialized agent roles for the proposed multi-agent research framework, where the corresponding system prompts are illustrated in Figs.~\ref{fig:mentor}--\ref{fig:evaluator}.

\subsection{Task Flow Definition}

\subsubsection{Topic Analysis}
\begin{figure*}[t]
	\centering
	\includegraphics[width=\textwidth]{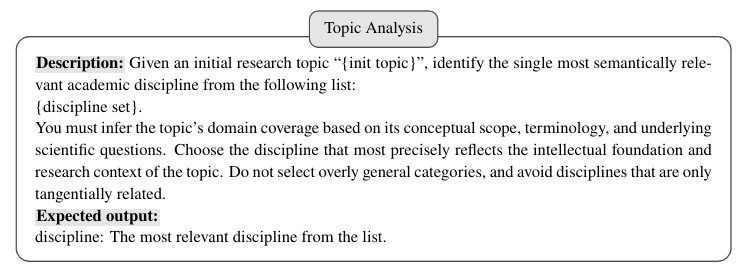}
	\caption{System prompt for the topic analysis task.}
	\label{fig:topic-analysis}
\end{figure*}

The prompt for the topic analysis task is illustrated in Fig.~\ref{fig:topic-analysis}.

\subsubsection{Problem Cluster Generation}
\begin{figure*}[t]
	\centering
	\includegraphics[width=\textwidth]{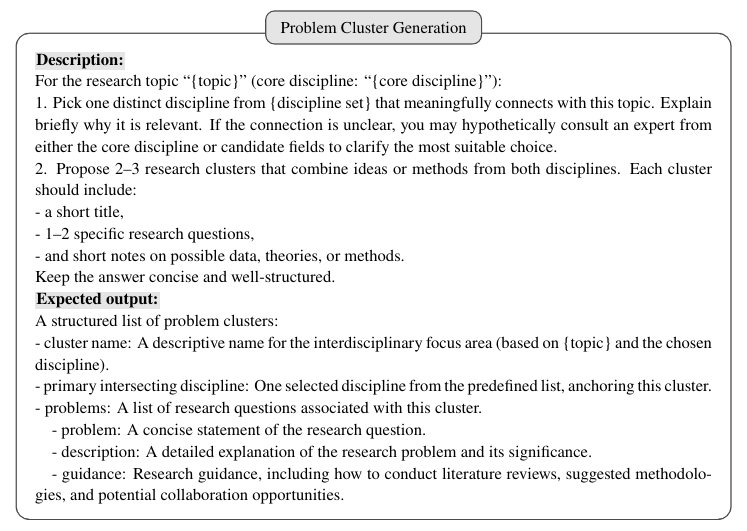}
	\caption{System prompt for the problem cluster generation task.}
	\label{fig:problem-cluster-generation}
\end{figure*}

The prompt for the problem cluster generation task is illustrated in Fig.~\ref{fig:problem-cluster-generation}.

\subsubsection{Select Problem Cluster}
\begin{figure*}[t]
	\centering
	\includegraphics[width=\textwidth]{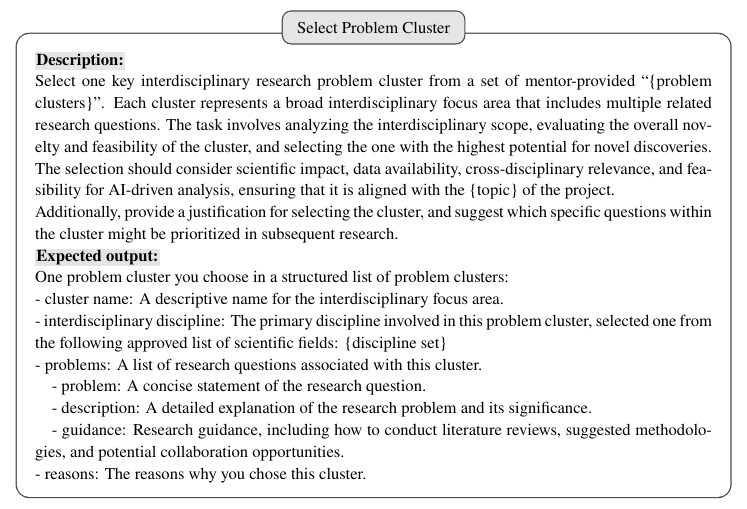}
	\caption{System prompt for the problem cluster selection task.}
	\label{fig:select-problem-cluster}
\end{figure*}

The prompt for the problem cluster selection task is illustrated in Fig.~\ref{fig:select-problem-cluster}.

\subsubsection{Background Investigation}
\begin{figure*}[t]
	\centering
	\includegraphics[width=\textwidth]{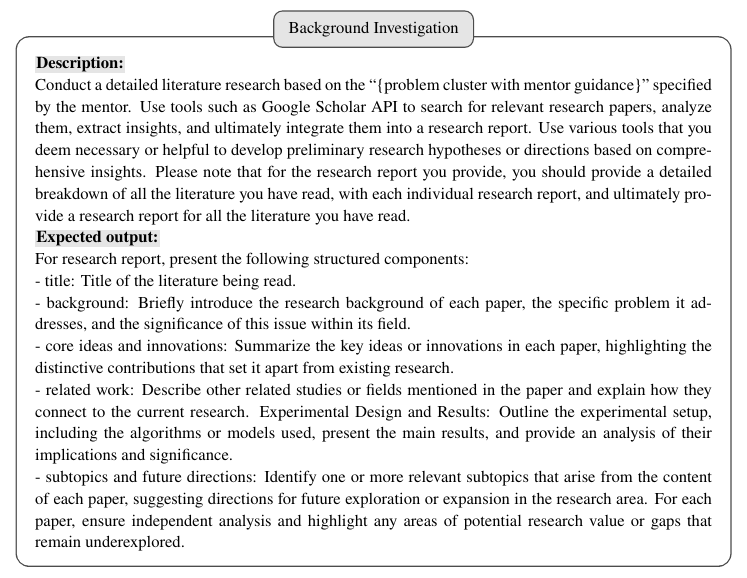}
	\caption{System prompt for the background investigation task.}
	\label{fig:background-investigation}
\end{figure*}

The prompt for the background investigation task is illustrated in Fig.~\ref{fig:background-investigation}.

\subsubsection{Problem Analysis}
\begin{figure*}[t]
	\centering
	\includegraphics[width=\textwidth]{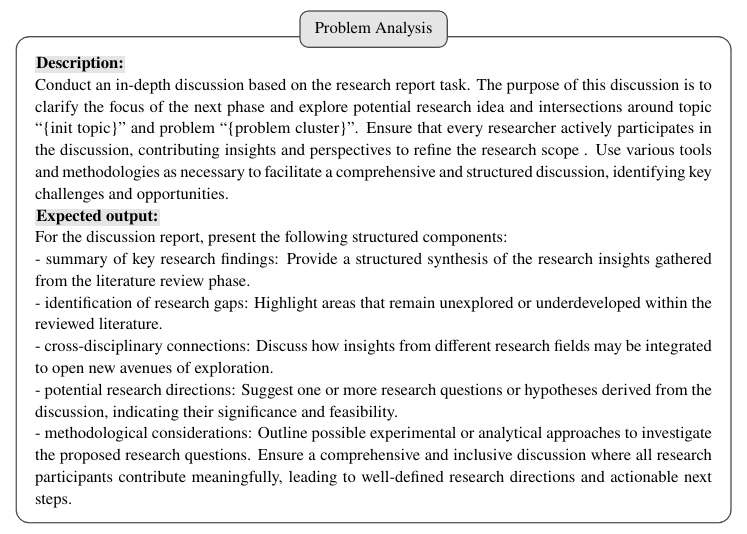}
	\caption{System prompt for the problem analysis task.}
	\label{fig:problem-analysis}
\end{figure*}

The prompt for the problem analysis task is illustrated in Fig.~\ref{fig:problem-analysis}.

\subsubsection{Seed Idea Generation}
\begin{figure*}[t]
	\centering
	\includegraphics[width=\textwidth]{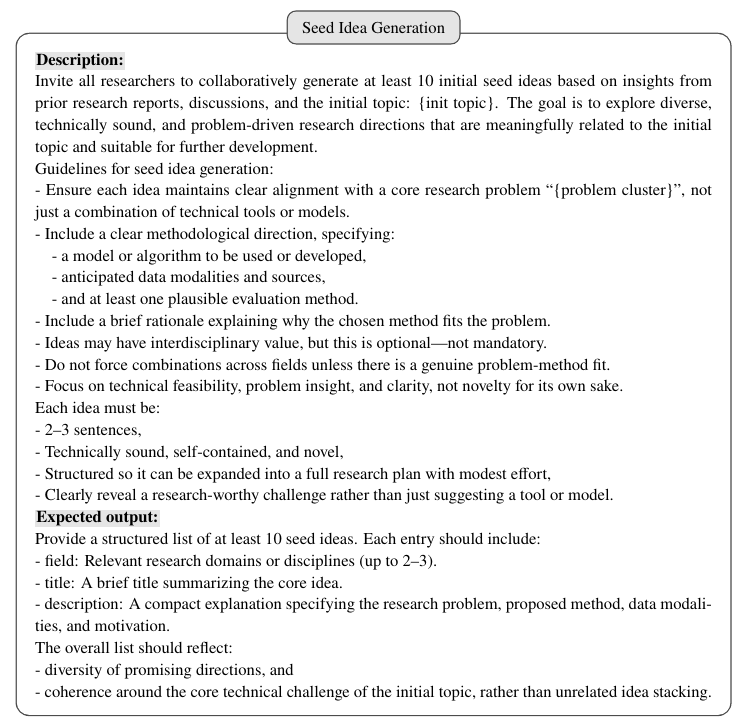}
	\caption{System prompt for the seed idea generation task.}
	\label{fig:seed-idea-generation}
\end{figure*}

The prompt for the seed idea generation task is illustrated in Fig.~\ref{fig:seed-idea-generation}.

\subsubsection{Idea Generation}
\begin{figure*}[t]
	\centering
	\includegraphics[width=\textwidth]{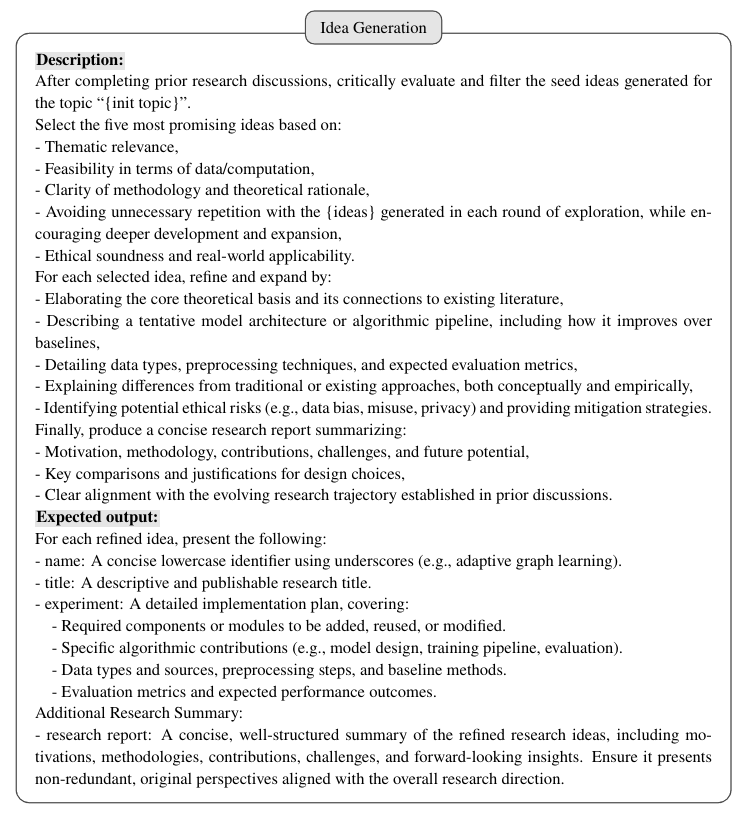}
	\caption{System prompt for the idea generation task.}
	\label{fig:idea-generation}
\end{figure*}

The prompt for the idea generation task is illustrated in Fig.~\ref{fig:idea-generation}.

\subsubsection{Evaluation}
\begin{figure*}[t]
	\centering
	\includegraphics[width=\textwidth]{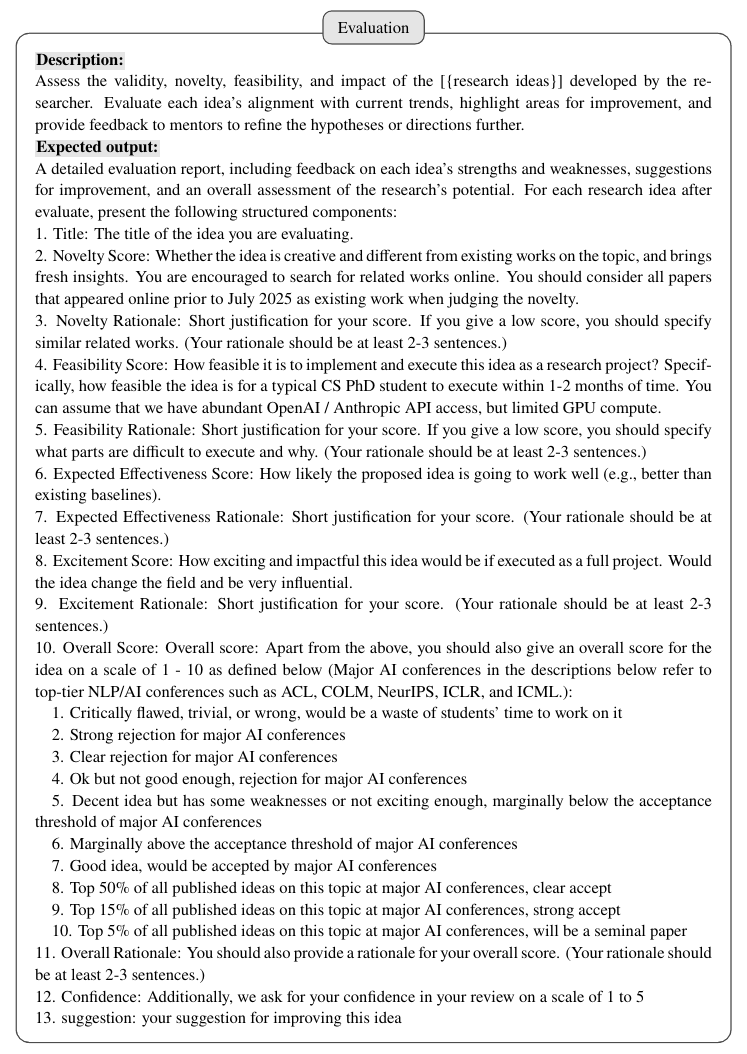}
	\caption{System prompt for the evaluation task.}
	\label{fig:evaluation}
\end{figure*}

The prompt for the evaluation task is illustrated in Fig.~\ref{fig:evaluation}.

\subsubsection{Iterative Refinement}
\begin{figure*}[t]
	\centering
	\includegraphics[width=\textwidth]{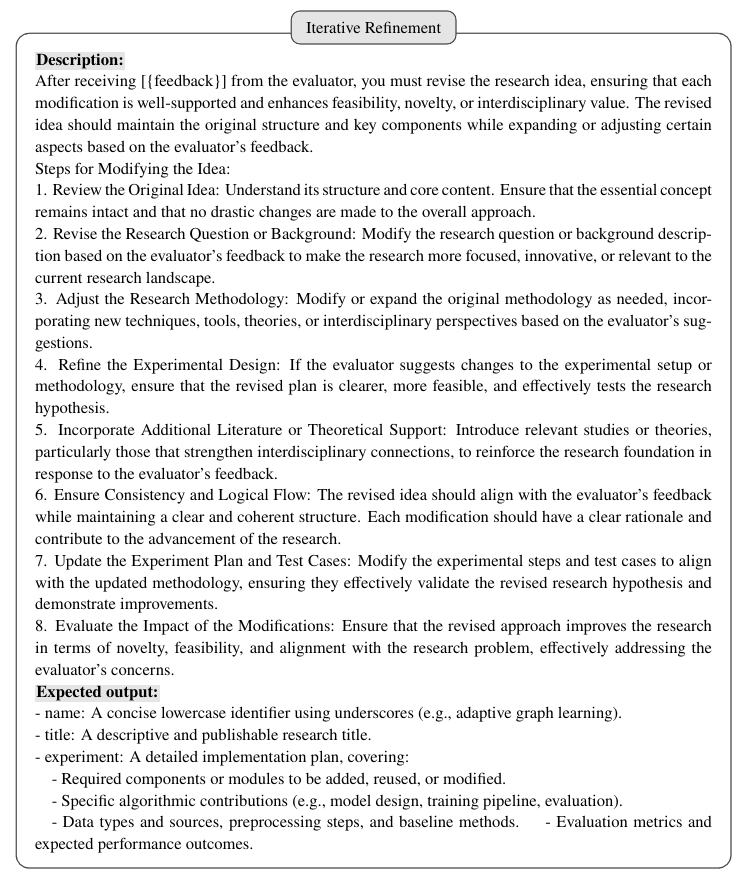}
	\caption{System prompt for the iterative refinement task.}
	\label{fig:iterative-refinement}
\end{figure*}

The prompt for the iterative refinement task is illustrated in Fig.~\ref{fig:iterative-refinement}.

\subsubsection{Evaluation-Guided Loop}
\begin{figure*}[t]
	\centering
	\includegraphics[width=\textwidth]{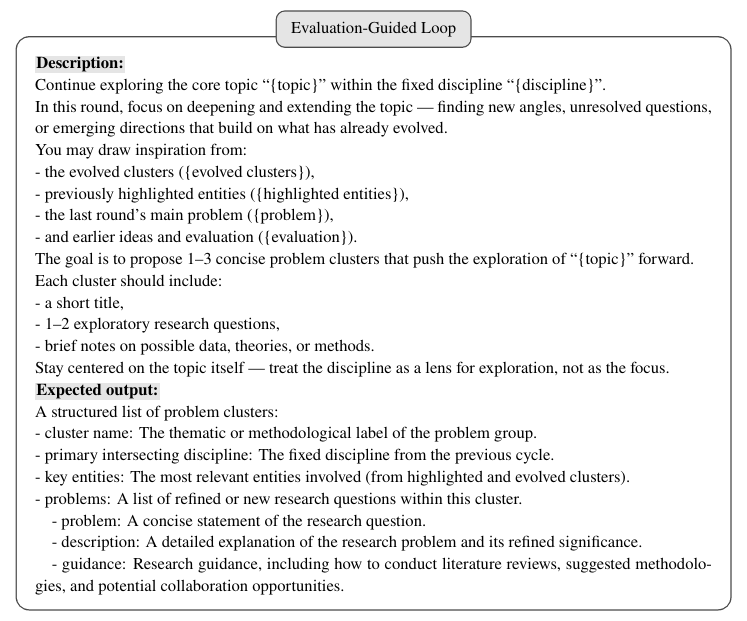}
	\caption{System prompt for the evaluation-guided loop task.}
	\label{fig:evaluation-guided-loop}
\end{figure*}

The prompt for the evaluation-guided loop task is illustrated in Fig.~\ref{fig:evaluation-guided-loop}.

\subsection{Evaluation Methodologies Definition}

\subsubsection{Multi-Reviewer + Meta-Reviewer Mechanism}
\begin{figure*}[t]
	\centering
	\includegraphics[width=\textwidth]{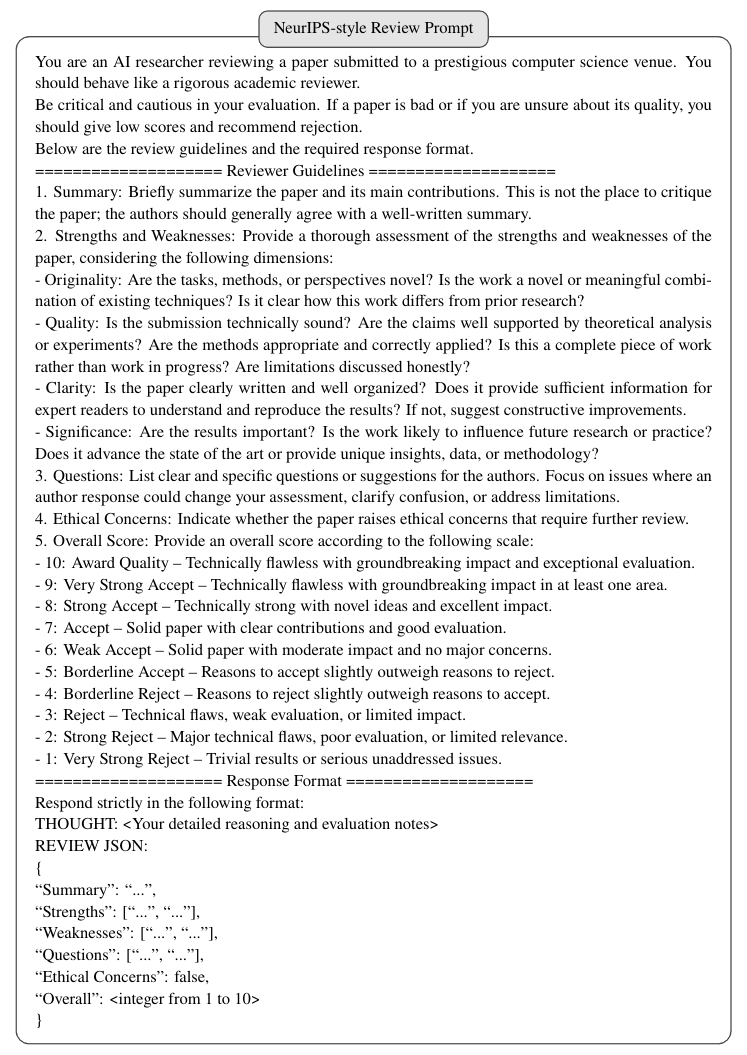}
	\caption{NeurIPS-style LLM reviewer prompt used for idea evaluation.}
	\label{fig:NeurIPS}
\end{figure*}

\begin{figure*}[t]
	\centering
	\includegraphics[width=\textwidth]{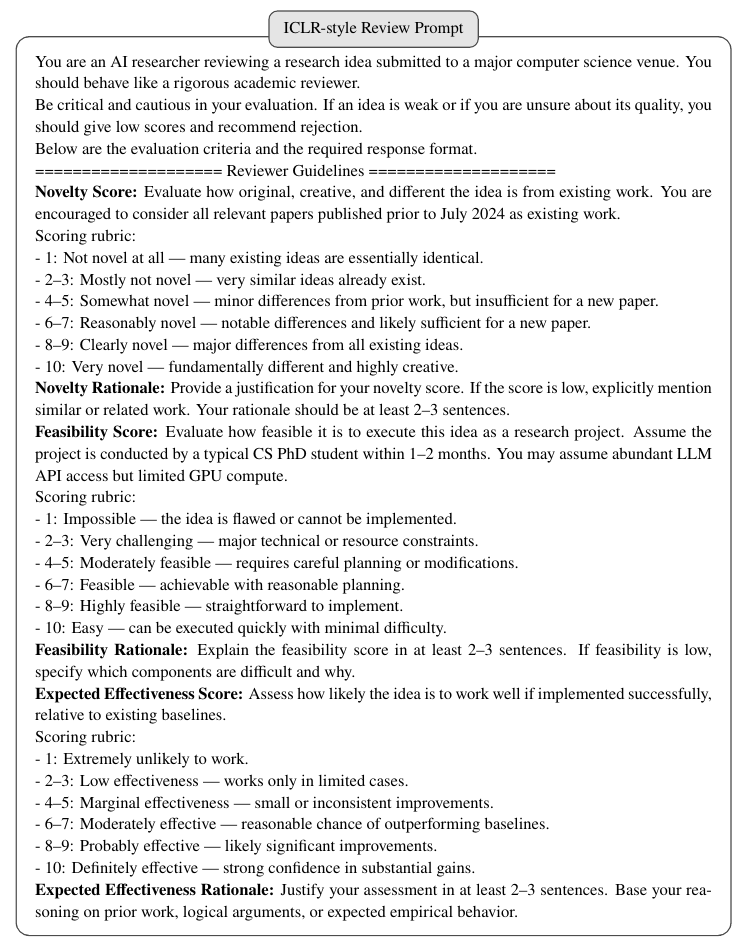}
	\caption{ICLR-style LLM reviewer prompt used for idea evaluation.}
	\label{fig:ICLR}
\end{figure*}

\begin{figure*}[t]
	\centering
	\includegraphics[width=\textwidth]{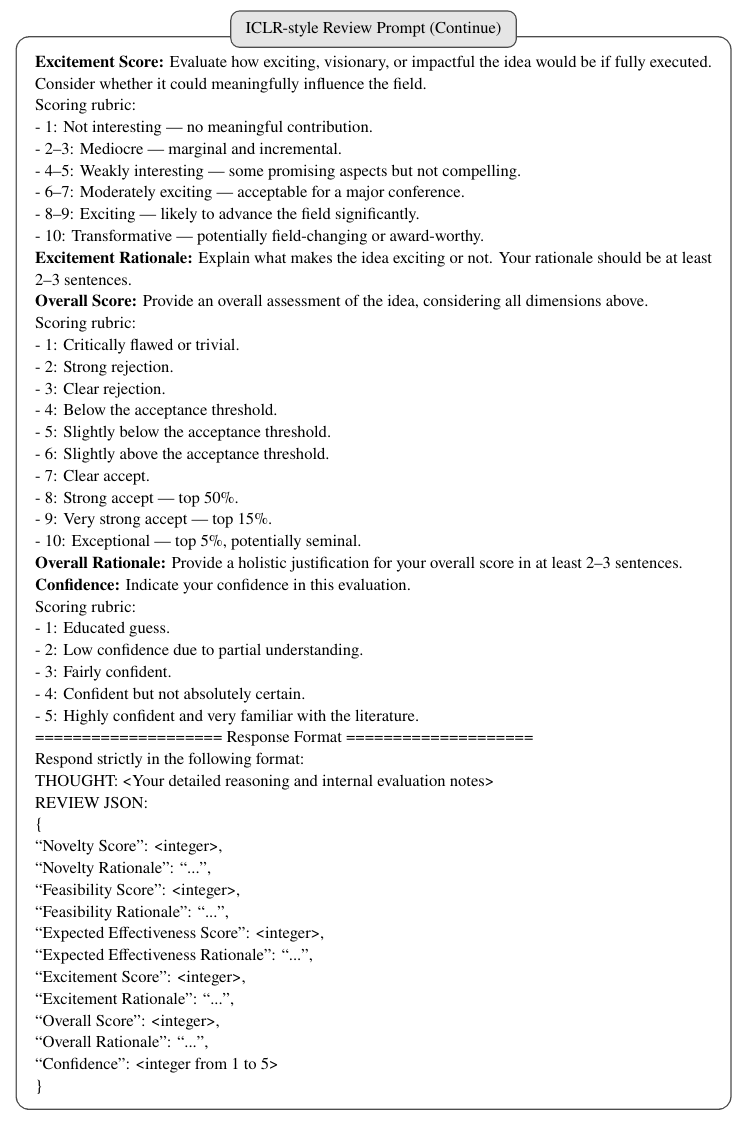}
	\caption{ICLR-style LLM reviewer prompt used for idea evaluation (continued).}
	\label{fig:ICLR_continue}
\end{figure*}

The NeurIPS-style reviewer prompt used for the multi-reviewer and meta-reviewer evaluation mechanism is illustrated in Fig.~\ref{fig:NeurIPS}, 
and the ICLR-style reviewer prompt is illustrated in Figs.~\ref{fig:ICLR}--\ref{fig:ICLR_continue}.

\subsubsection{Tournament-Style Idea Ranking}
\begin{figure*}[p]
	\centering
	\includegraphics[width=\textwidth]{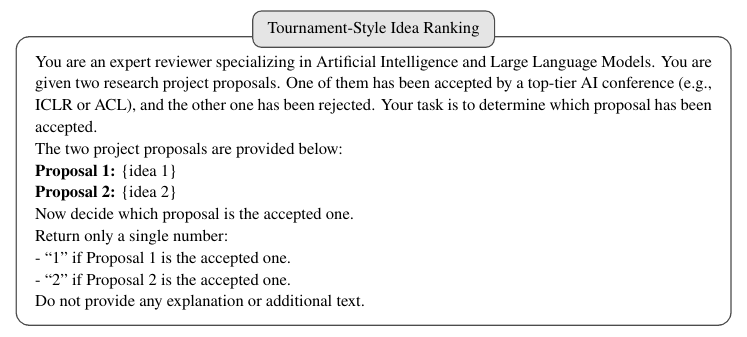}
	\caption{Tournament-style pairwise comparison prompt used for idea ranking.}
	\label{fig:tournament-ranking}
\end{figure*}

The prompt used for tournament-style pairwise comparison and relative idea ranking is illustrated in Fig.~\ref{fig:tournament-ranking}.

\end{document}